\documentclass[11pt]{article} 

\usepackage{fullpage}

\usepackage{comment}
\usepackage{url}
\usepackage{amsmath}
\usepackage{graphicx,xspace}

\usepackage{microtype}
\usepackage{booktabs} 
\usepackage{amsfonts}       
\usepackage{nicefrac}       
\usepackage{microtype}      

\usepackage{caption}
\usepackage{subcaption}
\usepackage{color}
\usepackage{mdwlist}

\usepackage{paralist}
\usepackage{algorithm,algorithmic}
\newfloat{Algorithm}{tb}{lox}
\floatname{Algorithm}{Algorithm}

\usepackage[colorlinks,linkcolor=red,citecolor=blue,urlcolor=blue]{hyperref}
\usepackage{caption}
\usepackage{subcaption}
\DeclareCaptionType{copyrightbox}
\usepackage{tabularx}
\usepackage{etoolbox}
\usepackage{mdwlist}
\usepackage{breakurl}

\usepackage[numbers,sort&compress]{natbib}

\usepackage{graphicx} 
\usepackage{tikz}
\usetikzlibrary{positioning,patterns,shapes, shadows}
\usepackage{pgfplots}
\usepgfplotslibrary{external}
\usepackage{multirow}
\usepackage{rotating}
\usepackage{wrapfig}
\usepackage{array} 
\usepackage{pbox}

\colorlet{darkgreen}{green!50!black}
\colorlet{darkred}{red!90!black}
\colorlet{darkyellow}{yellow!90!black}

\definecolor{blue-green}{rgb}{0.0, 0.87, 0.87}
\definecolor{brightturquoise}{rgb}{0.03, 0.91, 0.87}
\definecolor{darkturquoise}{rgb}{0.0, 0.81, 0.82}

\tikzset{
  gnode/.style={
    fill=white,
    draw=black,
    circle,
    very thick, 
    inner sep=3.5,
    drop shadow={shadow xshift=0.3ex,shadow yshift=-0.5ex, path
      fading={circle with fuzzy edge 20 percent}}
  }
}

\tikzset{
  rnode/.style={
    fill=black,
    draw=black,
    circle,
    very thick, 
    inner sep=3.5,
    drop shadow={shadow xshift=0.3ex,shadow yshift=-0.5ex, path
      fading={circle with fuzzy edge 20 percent}}
  }
}

\tikzset{
  ynode/.style={
    fill=black!50!white,
    draw=black,
    circle,
    very thick, 
    inner sep=3.5,
    drop shadow={shadow xshift=0.3ex,shadow yshift=-0.5ex, path
      fading={circle with fuzzy edge 20 percent}}
  }
}

\usepackage{amssymb}
\usepackage{pifont}
\usepackage{natbib}

\usepackage{algorithm}
\usepackage{algorithmic}

\input{./Definitions}
\graphicspath {{./Figures/}}

\begin{document}

\def\gridwidth{2}
  \def\Ngridwidth{3.5}
  \def\nodenum{4}
  \def\asnwidth{0.5}
  \def\asnmargin{0.05}
  \def\Xdim{13}

\newcommand{\asnbox}[3]{
  \fill [#3!25!lightgray]
  (#1 * \asnwidth + \asnmargin -\gridwidth * \nodenum,
  \Ngridwidth * 3 - #2 * \Ngridwidth + \asnmargin)
  rectangle
  (#1 * \asnwidth + \asnwidth - \asnmargin -\gridwidth * \nodenum,
  \Ngridwidth * 3 - #2 * \Ngridwidth + \Ngridwidth - \asnmargin);
  \draw
  (#1 * \asnwidth + \asnmargin -\gridwidth * \nodenum,
  \Ngridwidth * 3 - #2 * \Ngridwidth + \asnmargin)
  rectangle
  (#1 * \asnwidth + \asnwidth - \asnmargin -\gridwidth * \nodenum,
  \Ngridwidth * 3 - #2 * \Ngridwidth + \Ngridwidth - \asnmargin);
}

  \newcommand{\dasnbox}[2]{
    \draw[dashed]
    (#1 * \asnwidth + \asnmargin -\gridwidth * \nodenum,
    \Ngridwidth * 3 - #2 * \Ngridwidth + \asnmargin)
    rectangle
    (#1 * \asnwidth + \asnwidth - \asnmargin -\gridwidth * \nodenum,
    \Ngridwidth * 3 - #2 * \Ngridwidth + \Ngridwidth - \asnmargin);
  }

  \newcommand{\mvarr}[4]{
    \draw [->]
    (#1 * \asnwidth + 0.5 * \asnwidth -\gridwidth * \nodenum,
    3 * \Ngridwidth - #2 * \Ngridwidth + 0.5 * \Ngridwidth)
    --
    (#3 * \asnwidth + 0.5 * \asnwidth -\gridwidth * \nodenum,
    3 * \Ngridwidth - #4 * \Ngridwidth + 0.5 * \Ngridwidth);
  }

  \newcommand{\boxlayout}{
    \draw (-\gridwidth * \nodenum, 0)
    rectangle (0, \Ngridwidth * \nodenum);

    \fill [red!10] (-\gridwidth * \nodenum, 3 * \Ngridwidth)
    rectangle (0, 4 * \Ngridwidth);

    \fill [green!10] (-\gridwidth * \nodenum, 2 * \Ngridwidth)
    rectangle (0, 3 * \Ngridwidth);

    \fill [blue!10] (-\gridwidth * \nodenum, 1 * \Ngridwidth)
    rectangle (0, 2 * \Ngridwidth);

    \fill [brown!10] (-\gridwidth * \nodenum,0)
    rectangle (0, 1 * \Ngridwidth);

    \foreach \y in {1,2,3}
    {
      \draw[densely dashed]
      (-\gridwidth * \nodenum -\gridwidth * 0.1, \Ngridwidth * \y)
      --
      ( \Ngridwidth * 0.1,  \Ngridwidth * \y);
    }

  }

  \newcommand{\ptrna}{crosshatch}
  \newcommand{\ptrnb}{north west lines}
  \newcommand{\ptrnc}{north east lines}
  \newcommand{\ptrnd}{horizontal lines}

  \newcommand{\ptcla}{red}
  \newcommand{\ptclb}{green}
  \newcommand{\ptclc}{blue}
  \newcommand{\ptcld}{brown}

  \newcommand{\drawXpart}{
    \draw[pattern=\ptrnd, pattern color=\ptcld]
    ( \gridwidth * 0.5, 0 * \Ngridwidth)
    rectangle
    (\gridwidth * 0.5 + \Xdim,
    0 * \Ngridwidth + 1 * \Ngridwidth);

    \draw[pattern=\ptrnc, pattern color=\ptclc]
    (\gridwidth * 0.5,
    1 * \Ngridwidth)
    rectangle
    (\gridwidth * .5 + \Xdim,
    1 * \Ngridwidth + 1 * \Ngridwidth);

    \draw[pattern=\ptrnb, pattern color=\ptclb]
    (\gridwidth * 0.5,
    2 * \Ngridwidth)
    rectangle
    (\gridwidth * 0.5 + \Xdim,
    2 * \Ngridwidth + 1 * \Ngridwidth);

    \draw[pattern=\ptrna, pattern color=\ptcla]
    (\gridwidth * 0.5,
    3 * \Ngridwidth)
    rectangle
    (\gridwidth * 0.5 + \Xdim,
    3 * \Ngridwidth + 1 * \Ngridwidth);
  }

  \newcommand{\drawThetapart}[4]{
    \draw[pattern=#3, pattern color=#4]
    ( \gridwidth * 0.5,
    \nodenum * \Ngridwidth + \nodenum * \gridwidth + 0.25 * \Ngridwidth - #1 * \asnwidth + \asnwidth)
    rectangle
    ( \Xdim + \gridwidth * 0.5,
    \nodenum * \Ngridwidth + \nodenum * \gridwidth + 0.25 * \Ngridwidth - #2 * \asnwidth);

    \draw[pattern=#3, pattern color=#4]
    ( -\nodenum * \gridwidth + #1 * \asnwidth ,
    \nodenum * \Ngridwidth + 0.25 * \Ngridwidth)
    rectangle
    ( -\nodenum * \gridwidth + #2 * \asnwidth + \asnwidth,
    \nodenum * \Ngridwidth + 0.5 * \Ngridwidth );
  }

  \newcommand{\drawLable}{
    \node at (0.5 * \gridwidth + 0.5*\Xdim , 0.625*\nodenum*\Ngridwidth) {$x$};
    \node at ( 0 , \nodenum * \Ngridwidth + 
    0.6*\gridwidth*\nodenum) {$\varitheta$};
    \node at (- 1.1*\gridwidth * \nodenum, 0.5*\Ngridwidth * \nodenum ) {$\variz$};
    \node at (-0.5 *\gridwidth * \nodenum, \Ngridwidth*\nodenum + 
    0.8*\Ngridwidth ) {$\varipi$};
  }


\newcommand*\samethanks[1][\value{footnote}]{\footnotemark[#1]}

\title{Extreme Stochastic Variational Inference: Distributed and Asynchronous}
\author{
Jiong Zhang\thanks{contributed equally}\\
       {University of Texas, Austin}\\
       {zhangjiong724@utexas.edu}
\and
Parameswaran Raman\samethanks\\
       {University of California, Santa Cruz}\\
       {params@ucsc.edu}
\and
Shihao Ji\\
        {Georgia State University, Atlanta}\\
        {sji@cs.gsu.edu}
\and
Hsiang-Fu Yu\\
        {Amazon}\\
        {rofu.yu@gmail.com}        
\and
S.V.N Vishwanathan\\
        {University of California, Santa Cruz}\\
        {vishy@ucsc.edu}
\and
Inderjit S. Dhillon\\
        {University of Texas, Austin}\\
        {inderjit@cs.utexas.edu}        
}

\maketitle

\begin{abstract}
  Stochastic variational inference
  (SVI), the state-of-the-art algorithm for scaling variational
  inference to large-datasets, is inherently serial. Moreover, it
  requires the parameters to fit in the memory of a single processor;
  this is problematic when the number of parameters is in billions. In this paper, we 
  propose extreme stochastic variational inference (ESVI), an
  asynchronous and lock-free algorithm to perform variational inference
  for mixture models on massive real world datasets. ESVI
  overcomes the limitations of SVI by requiring that each processor only
  access a subset of the data and a subset of the parameters, thus
  providing data and model parallelism simultaneously.  We demonstrate
  the effectiveness of ESVI by running Latent Dirichlet Allocation (LDA)
  on UMBC-3B, a dataset that has a vocabulary of 3 million and a token size of 
  3 billion. In our experiments, we found that ESVI not only outperforms VI and SVI in wallclock-time, but also
  achieves a better quality solution.  In addition, we propose a
  strategy to speed up computation and save memory when fitting large
  number of topics.
\end{abstract}


\section{Introduction}
\label{sec:Introduction}

In recent years, variational inference (VI) has emerged as a powerful
technique for parameter estimation in a wide variety of Bayesian models
\cite{WaiJor08}, \cite{BleKucMca16}. One attractive property of VI is
that it reduces parameter estimation to the task of optimizing a
objective function, often with a well defined ``structure''. This opens
up the possibility of bringing to bear mature tools from optimization to
tackle massive problems. In this paper, we will primarily focus on {\it mixture models}, a large and
important class of latent variable models in machine learning which involve {\it local and global variables}. 
Traditionally, VI in mixture models involves alternating between updating global variables
and local variables. Both these operations involve accessing all the
data points.  Large datasets are usually stored on disk, and the cost of
accessing every datapoint to perform updates is prohibitively
high. Consequently, application of Bayesian methods was limited to small
and medium sized datasets.

In the literature, there are two main approaches to tackle the above
problem. The first is to use a divide and conquer strategy to distribute
the computation, and the second is to exploit the underlying structure
of the optimization problem to reduce the number of iterations (and
therefore the corresponding data access). For instance, one can divide
the data across multiple machines and use a map-reduce based framework
to aggregate the computations. See, for instance, \citep{NeiWanXin15}
for an example of this approach. An instance of the algorithmic approach
is given in \citep{HofBleWanPai13}. The key observation here is that the
optimization problem corresponding to the local variables is
\emph{separable}, that is, it can be written as a sum of functions,
where each function only depends on one data point. Therefore, one can
use stochastic optimization to update the local variables. Moreover, in
the Stochastic Variational Inference (SVI) algorithm of
\citep{HofBleWanPai13}, even before one pass through the dataset, the
global variables are updated multiple times, and therefore the model
parameters converge rapidly towards their final values. The argument is
similar in spirit to how stochastic optimization outperforms batch
algorithms for maximum aposteriori (MAP) estimation
\cite{BotBou11}. Consequently, SVI enabled applying variational
inference to datasets with millions of documents such as \textit{Nature}
and \textit{NewYork Times} \cite{HofBleWanPai13}, which could not be
handled before.



With the advent of the big-data era, we now routinely deal with
industry-scale problems involving billions of documents and tokens. Such
massive datasets pose another challenge, which, unfortunately 
VI and SVI are unable to address; namely, the set of parameters is so large
that {\it all the parameters do not fit} on a single processor\footnote{The
  discussion in this paper applies to the shared memory, distributed
  memory, as well as hybrid settings, and therefore we will use the term
  processor to denote either a thread or a machine.}. For instance, if
we have $D$ dimensional data and $K$ mixture components, then the
parameter size is $O\rbr{D \times K}$. If $D$ is of the order of
millions and $K$ is in the 100's or 1000's, modest numbers by todays
standards, the parameter size is a few 100s of GB (see our experiments
in Section \ref{sec:expmts}). 

In this paper we propose a new framework, Extreme Stochastic Variational
Inference (ESVI) to overcome these storage limitations and perform variational inference on datasets which are an order of 
magnitude larger than those that can be handled by any existing algorithm.
The main contributions of this paper are:
\begin{enumerate}
\item In ESVI, we develop a novel approach to achieve {\it simultaneous data and model parallelism} in {\it mixture models} by exploiting the following key idea: instead of
updating all the $K$ coordinates of a local variable and then updating
all $K$ global variables, we only update a small subset of the local
variables and the corresponding global variables.  The global variables
nomadically move through the network across workers, and this ensures mixing (see
Section \ref{sec:Parallelization} for more details). This seemingly
simple idea has some powerful consequences. It allows multiple
processors to simultaneously perform parameter updates.
\item Using a classic owner-computes paradigm, we make ESVI {\it asynchronous} and {\it lock-free}, and thus avoid expensive bulk synchronization between processors. This provides significant speedups in the multi-core multi-machine setting.
\item We present an extensive empirical study to evaluate the performance of ESVI by applying it to GMM and LDA models on several large real-world datasets.  We observe that ESVI outperforms VI and SVI both in terms of time as well as the quality of solutions obtained. In addition, we develop a variant ESVI-LDA-TOPK to speed up 
computation and save memory when fitting large number of topics. Section \ref{sec:expmts} describes our experiments and Appendix \ref{sec:varying_C_topk} discusses the TOPK strategy in more detail.
\end{enumerate}

To the best of our knowledge there is no existing algorithm
for VI that sports these desirable properties. Although, in principle,
ESVI is applicable even when data and/or model parameters fit in memory,
it truly shines for massive datasets where both model and data parallelism
are essential.


The rest of the paper is structured as follows: We discuss related work in 
Section \ref{sec:RelatedWork}. We briefly review
VI and SVI in Section \ref{sec:VariInfer}. We present our new algorithm ESVI in Section
\ref{sec:ExtrStochVari}, and discuss its advantages. Empirical evaluation is
presented in Section \ref{sec:expmts}, and Section \ref{sec:Conclusion} concludes the paper.


\section{Related Work}
\label{sec:RelatedWork}
\vspace{-0.5em}
Recent research on variational inference has focused on extending variational inference 
to non-conjugate models \citep{WanBle14} and developing variants that can scale to large datasets such as Stochastic Variational
Inference (SVI) \citep{HofBleWanPai13}. Other than the fact that SVI is inherently serial, it
also suffers from another drawback: storage of the entire $D \times K$
matrix $\theta$ on a single machine. On the other hand, our method, ESVI, 
exhibits model parallelism; each processor only needs to store $1/P$ fraction of $\theta$. 
Black-box variational inference (BBVI) \citep{RanGerBle14} generalizes SVI beyond conditionally conjugate models. The paper proposes a more generic framework by observing that the expectation in the ELBO can
be exploited directly to perform stochastic optimization. We view this
line of work as complementary to our research. It would be interesting
to verify if an ESVI like scheme can also be applied to BBVI. 

There has been a flurry of work in the past few years in developing data-parallel distributed methods for Approximate Bayesian Inference. One such popular work includes a classic Map-Reduce style inference algorithm \citep{NeiWanXin15}, where the data is divided across several worker nodes and
each of them perform VI updates in parallel until a final
synchronization step during which the parameters from the slaves are
combined to produce the final result. This method suffers from the
well-known \emph{curse of the last reducer}, that is, a single slow
machine can dramatically slow down the performance. {\it ESVI does not suffer from this 
problem}, because our asynchronous and lock-free updates avoid bulk synchronization altogether. 

\citep{BroBoyWibWilJor13} presents an algorithm that applies VI to the streaming setting by
performing asynchronous Bayesian updates to the posterior as batches of
data arrive continuously, which is similar in spirit to Hogwild \citep{RecReWriNiu11}. Their approach uses a parameter 
server to enable asynchronous local updates. Unlike ESVI, their work cannot guarantee that - (a) each worker works on the latest parameters, 
(b) the global parameters are all parallely updated. In \citep{ArcErm15} the authors present {\it Incremental Variational Inference} which is also a distributed 
variational inference algorithm, however it is also only data-parallel. Besides, it requires tuning of a step-size and sequential access of global parameters. {\it ESVI avoids these drawbacks}.

A number of data-parallel approaches exist in the Exact Bayesian Inference literature as well. \citep{HasWebLieVolLakBluTeh17} is a distributed MCMC based approach where workers perform MCMC updates locally and these are aggregated by maintaining a posterior server. \citep{YuHsiYunVisDhi15} proposed a distributed asynchronous algorithm for parameter estimation in LDA \cite{BleNgJor03}. However, the algorithm is specialized to collapsed Gibbs sampling for LDA, and it is unclear how to extend it to other, more general, mixture models. {\it ESVI in contrast is a purely VI based method and provides model-parallelism in addition to data-parallelism}.

Somewhat close to our ESVI-TOPK approach is {\it Memoized Online Variational Inference for DP Mixture Models} \citep{HugSud13}. 
This paper describes the application of Expectation Truncation to mixture models. In their L-sparse method, unused dimensions are set to zero and used 
dimensions are shifted. In ESVI-TOPK, unused dimensions are averaged (1-sum of used dimensions).

Another related line of work is {\it Sparse EM} \citep{NeaHin98}. There are some high-level similarities to ESVI in that both the methods update a subset of latent variables 
at any given time while keeping others frozen. However there are some crucial differences: (a) Sparse EM is not a parallel algorithm while ESVI is, 
(b) Sparse EM needs to iterate between sparse EM update and full EM update (to select active dimensions occasionally) while each ESVI worker's job 
queue will continuously distribute Z's dimensions to ensure a good mixing, (c) Sparse EM selects active dimensions based on values of Z, 
while ESVI is designed to ensure the active dimensions of each worker is an unbiased sample of all dimensions.

Since the coordinate-ascent algorithm in VI can be formulated as a message
passing scheme applied to general graphical models, we believe ESVI is
also related to Variational Message Passing \citep{WinBis05}. This connection could be
made more concrete if we assume a Mixture Model setup in both cases. Both
the d-VMP algorithm (Algorithm 2 in \citep{MasMarLanNieSalRamMad17}) and ESVI de-couple the global
parameters to make the updates scalable, however they differ in some
fundamental aspects. d-VMP defines a disjoint partitioning of the
global parameters based on their markov-blankets. In contrast, ESVI
completely decentralizes the global parameter updates by requiring
that the local variables (or assignment vector $z_i$) need to only
satisfy local summation constraints (as discussed in Lemma 1 in
Section \ref{sec:ExtrStochVari}). As a side-note, the local updates in d-VMP algorithm do not
seem to be de-coupled across the mixture components, whereas this
holds true in the case of ESVI. This is partly the reason why {\it ESVI can
distribute the assignment matrix $Z$ in both dimensions (along $N$ as well
as $K$, where $N$: number of data points, $K$: number of mixture components)}.

Collapsed VI is a technique to marginalize some parameters either before or after applying the variational bound \citep{HenRatLaw12}. We believe 
ESVI can be applied here as well, as long as the model is an instance
of the mixture of exponential family. 

Automatic Differentiation for variational inference is another recent line of work \citep{AlpDusRajAndBle18} which attempts to transform the latent variables into a real coordinate space where differentiation is tractable. We believe ESVI can be applied to some of these models as well (e.g. Mean field ADVI) since the access pattern of variables (assuming the setup of mixture of exponential family models) follows the same structure as desired by ESVI.


\section{Parameter Estimation for Mixture of Exponential Families}
\label{sec:VariInfer}
In this paper we focus on parameter estimation for the
mixture of exponential families model which generalizes a 
wide collection of latent variable models such as 
\textit{latent dirichlet allocation (LDA)}, \textit{gaussian mixture models (GMM)}, 
and \textit{stochastic mixed membership block models}.

\subsection{Mixture Model}
\label{sec:MixtureModel}

Given $N$ observations $x = \cbr{x_{1}, \ldots, x_{N}}$, with each
$x_{i} \in \RR^{D}$, we wish to model $p\rbr{x}$ as a mixture of $K$
distributions from the exponential family. Let
$z = \cbr{z_{1}, \ldots, z_{N}}$, with each
$z_{i} \in \cbr{1, \ldots, K}$, denote \emph{local} latent variables;
intuitively, $z_{i}$'s denote which component the current data point was
drawn from. Moreover, let
$\theta = \cbr{\theta_{1}, \ldots, \theta_{K}}$ denote \emph{global}
latent variables; each $\theta_{k}$ represents the sufficient statistics
of an exponential family distribution. Finally, we denote the
$K$-dimensional simplex by $\Delta_{K}$, and let $\pi \in \Delta_{K}$ be
the mixing coefficients of the mixture model. Note that $\pi$ is also a
global latent variable. For instance, in a GMM the parameters
$\theta_{1:K}$ represent the mean and covariance of the $K$ Gaussians,
$\pi$ represents the mixing proportions, and $z_{1:N}$ is the
soft-assignment of a particular data point to one of the $K$ components.
The following data generation scheme underlies a mixture of exponential
family model:
\begin{align}
  \label{eq:pidist}
  p\rbr{\pi | \alpha} & = \text{Dirichlet}\rbr{\alpha}
\end{align}
For $k = 1, \ldots, K$
\begin{align}
  \label{eq:thetadist}
  p\rbr{\theta_{k} | n_{k}, \nu_{k}} & = \exp\rbr{\inner{n_{k} \cdot \nu_{k}}{\theta_{k}} - n_{k} \cdot g\rbr{\theta_{k}} - h\rbr{n_{k}, \nu_{k}}}
\end{align}
where, $n_{k}$ and $\nu_{k}$ are the parameters of the conjugate prior.\\
for $i = 1, \ldots, N$
\begin{align}
  \label{eq:zdist}
  p\rbr{z_{i} | \pi } & = \text{Multinomial}\rbr{\pi} \\
  \label{eq:xdist}
  p\rbr{x_{i} | z_{i}, \theta } & = \exp\rbr{\inner{\phi\rbr{x_{i}, z_{i}}}{\theta_{z_{i}}} - g\rbr{\theta_{z_{i}}}}
\end{align}
where, $\phi$ denotes the sufficient statistics.
Observe that $p\rbr{\theta_{k} | n_{k}, \nu_{k}}$ is conjugate to
$p\rbr{x_{i} | z_{i}=k, \theta_{k}}$, while $p\rbr{\pi | \alpha}$ is
conjugate to $p\rbr{z_{i} | \pi}$. The joint distribution of the data
and latent variables can be written as
\begin{align}
  \label{eq:joint_p}
  p\rbr{x, \pi, z, \theta | \alpha, n, \nu} = p\rbr{\pi | \alpha} \cdot \prod_{k=1}^{K} p\rbr{\theta_{k} | n_{k}, \nu_{k}} \cdot  
  \prod_{i=1}^{N} p\rbr{z_{i} | \pi} \cdot p\rbr{x_i | z_i, \theta}
\end{align}

\subsection{Variational Inference and Stochastic Variational Inference}
\label{sec:BatVariInfer}

The goal of inference is to estimate
$p\rbr{\pi, z, \theta | x, \alpha, n, \nu}$. However, computing this
distribution requires marginalization over $x$, which is typically
intractable. Therefore, variational inference \citep{BleKucMca16}
approximates this distribution with a fully, factorized distribution of
the following form:
\begin{align}
  \label{eq:joint_q}
  q\rbr{\pi, z, \theta | \varipi, \variz, \varitheta} = q\rbr{\pi | \varipi} \cdot \prod_{i=1}^{N} q\rbr{z_{i}| \variz_{i}} \cdot \prod_{k=1}^{K} q\rbr{\theta_{k} | \varitheta_{k}}.
\end{align}
{\it A $\sim$ over a symbol is used to denote that it is a parameter of the
variational distribution}. Note that $\variz_{i} \in \Delta_{K}$ and
$z_{i,k} = q\rbr{z_{i} = k | \variz_{i}}$. Moreover, each of the factors
in the variational distribution is assumed to belong to the same
exponential family as their full conditional counterparts in
\eqref{eq:joint_p}. The variational parameters are
estimated by maximizing the following evidence lower-bound (ELBO)
\citep{BleKucMca16}:
\begin{align}
	\label{eq:elbo1}
	\Lcal\rbr{\varipi, \variz, \varitheta} = \EE_{q\rbr{\pi, z, \theta | \varipi, \variz, \varitheta}}\sbr{\log p\rbr{x, \pi, z, \theta | \alpha, n, \nu}} 
	- \EE_{q\rbr{\pi, z, \theta | \varipi, \variz, \varitheta}}\sbr{\log q\rbr{\pi, z, \theta | \varipi, \variz, \varitheta}}
\end{align}

Variational inference algorithms perform coordinate ascent updates on
$\Lcal$ by optimizing each set of variables, one at a time.

\paragraph{Update for $\varipi$}
\begin{align}
  \label{eq:varipi_update}
  \varipi_{k} = \alpha + \sum_{i=1}^{N} \variz_{i,k}
\end{align}

\paragraph{Update for $\varitheta_{k}$}
The components of $\varitheta_{k}$ namely $\varin_{k}$ and $\varinu_{k}$
are updated as follows:
\begin{align}
  \label{eq:varin_update}
  \varin_{k} & = n_{k} + N_{k} \\
  \label{eq:varinu_update}
  \varinu_{k} & = n_{k} \cdot \nu_{k} + N_{k} \cdot \xbar_{k}
\end{align}
where $N_{k} := \sum_{i=1}^{N} \variz_{i,k}$ and $\xbar_{k} :=
\frac{1}{N_{k}} \sum_{i=1}^{N} \variz_{i,k} \cdot \phi\rbr{x_{i}, k}$.

\paragraph{Update for $\variz_{i}$}
Let $u_{i}$ be a $K$ dimensional vector whose $k$-th component is given
by
\begin{align}
  \label{eq:uk_def}
  u_{i,k} 
  & = \psi\rbr{\varipi_{k}} - \psi\rbr{\sum_{k'=1}^{K} \varipi_{k'}} 
  + \inner{\phi\rbr{x_{i}, k}}{\EE_{q\rbr{\theta_{k} |\varitheta_{k}}} \sbr{\theta_{k}}} - \EE_{q\rbr{\theta_{k} | \varitheta_{k}}} \sbr{g\rbr{\theta_{k}}} \\
  \label{eq:variz_update}
  \variz_{i, k} & = \frac{\exp\rbr{u_{i, k}}}{\sum_{k'=1}^{K} \exp\rbr{u_{i, k'}}}
\end{align}
where, $\psi\rbr{\cdot}$ denotes the digamma function, which is defined as the logarithmic derivative of the gamma function.
It has to be noted that the summation term $\psi\rbr{\sum_{k'=1}^{K} \varipi_{k'}}$ cancels out during the $\variz_{i,k}$ update in (\ref{eq:variz_update}). 
The VI algorithm \citep{WaiJor08, BleKucMca16} iteratively
performs the sequence of updates as illustrated in Algorithm \ref{alg:vi_pseudocode}.

\begin{algorithm}[H]
   \caption{VI}
   \label{alg:vi_pseudocode}
\begin{algorithmic}
\FOR {$i = 1, \ldots, N$}
  \STATE {Update $\variz_{i}$ using \eqref{eq:variz_update}}
\ENDFOR
\FOR {$k = 1, \ldots K$}
\STATE {Update $\varipi_k$ using \eqref{eq:varipi_update}}
\STATE {Update $\varitheta_{k}$ using \eqref{eq:varin_update} and \eqref{eq:varinu_update}}
\ENDFOR
\end{algorithmic}
\end{algorithm}
\begin{algorithm}[H]
   \caption{SVI}
   \label{alg:svi_pseudocode}
\begin{algorithmic}
\STATE {Generate step size sequence $\eta_t \in \rbr{0, 1}$}
\STATE {Pick an $i \in \cbr{1, \ldots, N}$ uniformly at random}
\begin{ALC@g}
\STATE {Update $\variz_{i}$ using \eqref{eq:variz_update}}
\FOR {$k = 1, \ldots K$}
\STATE {Update $\varipi_k\leftarrow(1-\eta_t)\varipi_k + \eta_t \rbr{\alpha+N\cdot\variz_{i,k}}$}
\STATE {$\hat{\varitheta}_{k}$=\{$n_k+N\cdot\variz_i,\>n_k\cdot\nu_k+N\cdot\variz_{i,k}\cdot\phi\rbr{x_i,k}$\}}
\STATE {Update $\varitheta_k\leftarrow(1-\eta_t)\varitheta_k + \eta_t \hat{\varitheta}_k$}
\ENDFOR
\end{ALC@g}
\end{algorithmic}
\end{algorithm}


The SVI algorithm \citep{HofBleWanPai13}, performs a 
slightly different sequence of updates
as shown in Algorithm \ref{alg:svi_pseudocode}. In contrast to VI where all the local variables are
updated before updating the global variables, here $\variz_{i}$
corresponding to one data point $x_{i}$ is updated, followed by updating the 
global parameters $\varipi$ and $\varitheta$.


\section{Extreme Stochastic Variational Inference (ESVI)}
\label{sec:ExtrStochVari}
In this paper we propose the following sequence of updates as illustrated in in Algorithm \ref{alg:esvi_pseudocode}.
\begin{algorithm}[H]
   \caption{ESVI}
   \label{alg:esvi_pseudocode}
\begin{algorithmic}
\STATE {Sample $i \in \cbr{1, \ldots, N}$}
\begin{ALC@g}
\STATE {Select $\Kcal \subset \cbr{1, \ldots, K}$}
\begin{ALC@g}
\STATE {Update $\variz_{i,k}$ for all $k \in \Kcal$ (see below)}
\STATE {Update $\varipi_{k}$ for all $k \in \Kcal$ using \eqref{eq:varipi_update}}
\STATE {Update $\varitheta_{k}$ for all $k \in \Kcal$ using \eqref{eq:varin_update}, \eqref{eq:varinu_update}}
\end{ALC@g}
\end{ALC@g}
\end{algorithmic}
\end{algorithm}

Before we discuss why this update is advantageous for parallelization,
let us first study how one can update a subset of coordinates of
$\variz_{i}$ efficiently. In order to demonstrate this, we first plug in the true joint distribution given by \eqref{eq:joint_p} and variational distribution given by \eqref{eq:joint_q} into
the ELBO \eqref{eq:elbo1}. Next, we restrict our attention to the terms in the ELBO which
depend on $z_{i}$ and substitute \eqref{eq:xdist},
$\variz_{i,k} = q\rbr{z_{i} = k | \variz_{i}}$ and
$\EE_{q\rbr{\pi | \varipi}}\sbr{\log p\rbr{z_{i} = k | \pi}} =
\psi\rbr{\varipi_{k}} - \psi\rbr{\sum_{k'=1}^{K} \varipi_{k'}}$.
This yields the following objective function,
\begin{align}
  \label{eq:elbo_zik}
  \Lcal \rbr{\variz_{i} | \varipi, \varitheta}
  & = \sum_{k=1}^{K} \variz_{i,k} \cdot \rbr{\psi\rbr{\varipi_{k}} -
    \psi\rbr{\sum_{k'=1}^{K} \varipi_{k'}}} \nonumber \\
  & + \sum_{k=1}^{K} \variz_{i,k} \cdot \rbr{\inner{\phi\rbr{x_{i}, k}}{\EE_{q\rbr{\theta_{k} |\varitheta_{k}}} \sbr{\theta_{k}}}
  - \EE_{q\rbr{\theta_{k} | \varitheta_{k}}} \sbr{g\rbr{\theta_{k}}} - \log \variz_{i,k}}.
\end{align}

Now using the definition of $u_{i,k}$ in \eqref{eq:uk_def}, one can
compactly rewrite the above objective function as
\begin{align}
  \label{eq:elbo_zik_compact}
  \Lcal \rbr{\variz_{i} | \varipi, \varitheta} = \sum_{k=1}^{K} \variz_{i,k} \cdot \rbr{u_{i,k} - \log \variz_{i,k}}.
\end{align}

Moreover, to ensure that $\variz_{i,k}$ is a valid distribution, one
needs to enforce the following constraints:
\begin{align}
  \label{eq:zi_sum_constraint}
  \sum_{k} \variz_{i, k} = 1, \quad \; 0 \leq \variz_{i, k} \leq 1.
\end{align}

The following lemma shows that one can find a closed form solution to
maximizing \eqref{eq:elbo_zik_compact} even if we restrict our attention
to a subset of variables.
\begin{lemma}
  \label{lemma:z-udpate}
  For $2 \leq K' \leq K$, let $\Kcal \subset \cbr{1, \ldots, K}$ be \; s.t. 
  $\abr{\Kcal} = K'$. For any $C > 0$, the problem
    \begin{align}
    \label{eq:z-update-general}
    \nonumber
      \max_{z_{i} \in \RR^{K'}} \;
      \; \Lcal_{\Kcal} & = \sum_{k\in \Kcal} \variz_{i,k} \cdot u_{i,k} - \variz_{i,k} \cdot \log \variz_{i,k} \; \; \; \; \; \\
      \text{s.t. } \;
      & \sum_{k\in \Kcal} \variz_{i,k} = C \quad \text{and} \quad 0 \leq \variz_{i,k},
    \end{align}
  has the closed form solution:
  \begin{align}
    \label{eq:optimal_z}
    \variz^{*}_{i,k} = C \frac{\exp\rbr{u_{i,k}}}{\sum_{k'\in \Kcal}
    \exp\rbr{u_{i,k'}}}, \text{ for } k \in \Kcal.
  \end{align}
\end{lemma}

\par\noindent{\bf Proof\ }
We prove that $\variz_{i}^*$ is a stationary point by checking the KKT
conditions for \eqref{eq:z-update-general}.  Let
$h\rbr{\variz_{i}} = \rbr{\sum_{k \in \Kcal} \variz_{i,k}}- C$ and
$g_{k}\rbr{\variz_{i}} = -z_{i,k}$.  It is clear that $\variz_{i}^*$
satisfies the primal feasibility. Now consider KKT multipliers:
  \begin{align*}
    \lambda = \log \frac{C}{\sum_{k'\in \Kcal} \exp\rbr{u_{i,k'}}} , \text{ and } \mu_{k} = 0.
  \end{align*}
  We have
  \begin{align*}
    \nabla_{k} \Lcal_{\Kcal}\rbr{\variz_{i}^{*}} &= u_{i,k}  - \log (\variz_{i,k}^*) - 1 \\
    & = u_{i,k} - \rbr{u_{i,k} + \log \frac{C}{\sum_{k'\in \Kcal} \exp(u_{i,k'})}} \\
    & = \log \frac{C}{\sum_{k'\in \Kcal} \exp(u_{i,k'})}\\
    \lambda \nabla_k h(\variz_{i}^*) &=  \log \frac{C}{\sum_{k'\in \Kcal}\exp(u_{i,k'})} \\
    \mu_k \nabla_k g_{k'}(\variz_{i}^*) & = 0.
  \end{align*}
  Then it is easy to verify that
  $\nabla_{k} \Lcal_{\Kcal} (\variz_{i}^*) = \lambda = \lambda \nabla_{k} h(\variz_{i}^*)$. Thus, $\variz_{i}^*$
  satisfies the stationarity condition:
  \[
    \nabla \Lcal_{\Kcal} (\variz_{i}^*) = \lambda \nabla h(\variz_{i}^*) + \sum_{k=1}^K \mu_k \nabla g_k(\variz_{i}^*).
  \]
  Due to choice of $\mu_k=0$, complementary slackness and dual feasibility are
  also satisfied. Thus, $\variz_{i}^*$ is the optimal solution to
  \eqref{eq:z-update-general}.
\hfill\BlackBox\\[2mm]

The lemma suggests the following strategy: start with a feasible
$\variz_{i}$, pick, say, a pair of coordinates $\variz_{i,k}$ and
$\variz_{i,k'}$ and let $\variz_{i,k} + \variz_{i,k'} = C$. Solve
\eqref{eq:z-update-general}, which has the closed form solution
\eqref{eq:optimal_z}. Clearly, if $\variz_{i}$ satisfied constraints
\eqref{eq:zi_sum_constraint} before the
update, it will continue to satisfy the constraints even after the
update. On the other hand, the conditional ELBO
\eqref{eq:elbo_zik_compact} increases as a result of the
update. Therefore, ESVI is a valid coordinate ascent algorithm for 
improving the ELBO \eqref{eq:elbo1}.

\subsection{Access Patterns}
\label{sec:AccessPatterns}

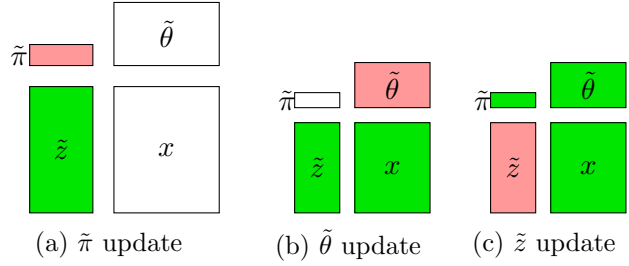
\begin{figure}[H]
  \centering
  \begin{subfigure}[t]{0.22\textwidth}
    \centering
    \begin{tikzpicture}[scale=0.28]

      \fill[green!90!black] (0,0) rectangle (3,6);
      \draw[black] (0,0) rectangle (3,6);
      \node at (1.5, 3) {$\variz$};
      \fill[white!40!white] (4,0) rectangle (9,6);
      \draw[black] (4,0) rectangle (9,6);
      \node at (6.5, 3) {$x$};
      \fill[red!40!white] (0,7) rectangle (3,8);
      \draw[black] (0,7) rectangle (3,8);
      \node at (-0.5, 7.5) {$\varipi$};
      \fill[white!40!white] (4,7) rectangle (9,10);
      \draw[black] (4,7) rectangle (9,10);
      \node at (6.5, 8.5) {$\varitheta$};
    \end{tikzpicture}
    \caption{$\varipi$ update}
  \end{subfigure}
  \begin{subfigure}[t]{0.15\textwidth}
    \centering
    \begin{tikzpicture}[scale=0.20]
      \fill[green!90!black] (0,0) rectangle (3,6);
      \draw[black] (0,0) rectangle (3,6);
      \node at (1.5, 3) {$\variz$};
      \fill[green!90!black] (4,0) rectangle (9,6);
      \draw[black] (4,0) rectangle (9,6);
      \node at (6.5, 3) {$x$};
      \fill[white!40!white] (0,7) rectangle (3,8);
      \draw[black] (0,7) rectangle (3,8);
      \node at (-0.5, 7.5) {$\varipi$};
      \fill[red!40!white] (4,7) rectangle (9,10);
      \draw[black] (4,7) rectangle (9,10);
      \node at (6.5, 8.5) {$\varitheta$};
    \end{tikzpicture}
    \caption{$\varitheta$ update}
  \end{subfigure}
   \begin{subfigure}[t]{0.15\textwidth}
    \centering
    \begin{tikzpicture}[scale=0.20]
      \fill[red!40!white] (0,0) rectangle (3,6);
      \draw[black] (0,0) rectangle (3,6);
      \node at (1.5, 3) {$\variz$};
      \fill[green!90!black] (4,0) rectangle (9,6);
      \draw[black] (4,0) rectangle (9,6);
      \node at (6.5, 3) {$x$};
      \fill[green!90!black] (0,7) rectangle (3,8);
      \draw[black] (0,7) rectangle (3,8);
      \node at (-0.5, 7.5) {$\varipi$};
      \fill[green!90!black] (4,7) rectangle (9,10);
      \draw[black] (4,7) rectangle (9,10);
      \node at (6.5, 8.5) {$\varitheta$};
    \end{tikzpicture}
    \caption{$\variz$ update}
  \end{subfigure}
  \vspace{1em}
  \caption{Access pattern of variables during Variational Inference (VI)
    updates. Green indicates that the variable or data point is being
    read, while red indicates that the variable is being updated.}
  \label{fig:accesspattern_vi}
\end{figure}

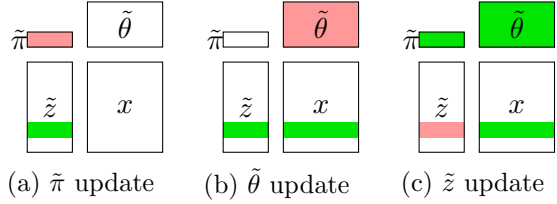
\begin{figure}[H]
  \centering
  \begin{subfigure}[t]{0.15\textwidth}
    \centering
    \begin{tikzpicture}[scale=0.20]
      \fill[white!40!white] (0,0) rectangle (3,6);
      \fill[green!90!black] (0,1) rectangle (3,2);
      \draw[black] (0,0) rectangle (3,6);
      \node at (1.5, 3) {$\variz$};
      \fill[white!40!white] (4,0) rectangle (9,6);
      \draw[black] (4,0) rectangle (9,6);
      \node at (6.5, 3) {$x$};
      \fill[red!40!white] (0,7) rectangle (3,8);
      \draw[black] (0,7) rectangle (3,8);
      \node at (-0.5, 7.5) {$\varipi$};
      \fill[white!40!white] (4,7) rectangle (9,10);
      \draw[black] (4,7) rectangle (9,10);
      \node at (6.5, 8.5) {$\varitheta$};
    \end{tikzpicture}
    \caption{$\varipi$ update}
  \end{subfigure}
  \begin{subfigure}[t]{0.15\textwidth}
    \centering
    \begin{tikzpicture}[scale=0.20]
      \fill[white!40!white] (0,0) rectangle (3,6);
      \fill[green!90!black] (0,1) rectangle (3,2);
      \draw[black] (0,0) rectangle (3,6);
      \node at (1.5, 3) {$\variz$};
      \fill[white!40!white] (4,0) rectangle (9,6);
      \fill[green!90!black] (4,1) rectangle (9,2);
      \draw[black] (4,0) rectangle (9,6);
      \node at (6.5, 3) {$x$};
      \fill[white!40!white] (0,7) rectangle (3,8);
      \draw[black] (0,7) rectangle (3,8);
      \node at (-0.5, 7.5) {$\varipi$};
      \fill[red!40!white] (4,7) rectangle (9,10);
      \draw[black] (4,7) rectangle (9,10);
      \node at (6.5, 8.5) {$\varitheta$};
    \end{tikzpicture}
    \caption{$\varitheta$ update}
  \end{subfigure}
   \begin{subfigure}[t]{0.15\textwidth}
    \centering
    \begin{tikzpicture}[scale=0.20]
      \fill[white!40!white] (0,0) rectangle (3,6);
      \fill[red!40!white] (0,1) rectangle (3,2);
      \draw[black] (0,0) rectangle (3,6);
      \node at (1.5, 3) {$\variz$};
      \fill[white!40!white] (4,0) rectangle (9,6);
      \fill[green!90!black] (4,1) rectangle (9,2);
      \draw[black] (4,0) rectangle (9,6);
      \node at (6.5, 3) {$x$};
      \fill[green!90!black] (0,7) rectangle (3,8);
      \draw[black] (0,7) rectangle (3,8);
      \node at (-0.5, 7.5) {$\varipi$};
      \fill[green!90!black] (4,7) rectangle (9,10);
      \draw[black] (4,7) rectangle (9,10);
      \node at (6.5, 8.5) {$\varitheta$};
    \end{tikzpicture}
    \caption{$\variz$ update}
  \end{subfigure}
  \vspace{1em}
  \caption{Access pattern of variables during Stochastic Variational
    Inference (SVI) updates. Green indicates that the variable or data
    point is being read, while red indicates that the variable is being
    updated.}
  \label{fig:accesspattern_svi}
\end{figure}

In this section we compare the access patterns of variables in the three algorithms 
to gain a better understanding of their abilities to be parallelized efficiently. In VI, the updates for $\varipi$ and $\varitheta$ requires access to all $\variz_{i}$, while update to $\variz_{i}$
requires access to $\varipi$ and all $\varitheta_{k}$. On the other hand, in case of SVI, the access pattern is somewhat different. The updates for $\varipi$ and $\varitheta$ require access to only the $\variz_{i}$ that was updated, however the
update to $\variz_{i}$ still requires access to $\varipi$ and all the $\varitheta_{k}$. This is a crucial bottleneck 
to model parallelism. Refer to Figure~\ref{fig:accesspattern_vi} and Figure~\ref{fig:accesspattern_svi} for a visual illustration. 

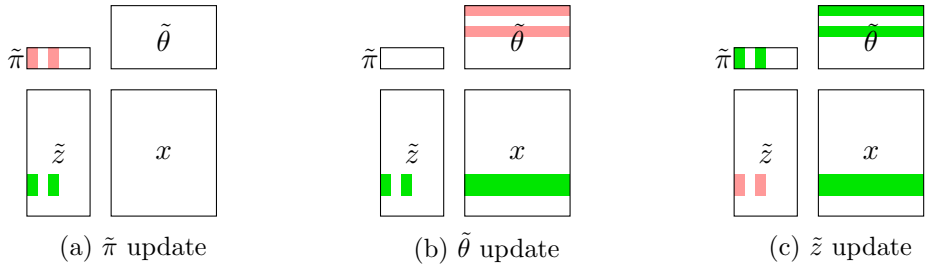
\begin{figure}[H]
  \centering
  \begin{subfigure}[t]{0.22\textwidth}
    \begin{tikzpicture}[scale=0.28]
      \fill[white!40!white] (0,0) rectangle (3,6);
      \fill[green!90!black] (0,1) rectangle (0.5,2);
      \fill[green!90!black] (1,1) rectangle (1.5,2);
      \draw[black] (0,0) rectangle (3,6);
      \node at (1.5, 3) {$\variz$};
      \fill[white!40!white] (4,0) rectangle (9,6);
      \draw[black] (4,0) rectangle (9,6);
      \node at (6.5, 3) {$x$};
      \fill[white!40!white] (0,7) rectangle (3,8);
      \fill[red!40!white] (0,7) rectangle (0.5,8);
      \fill[red!40!white] (1,7) rectangle (1.5,8);
      \draw[black] (0,7) rectangle (3,8);
      \node at (-0.5, 7.5) {$\varipi$};
      \fill[white!40!white] (4,7) rectangle (9,10);
      \draw[black] (4,7) rectangle (9,10);
      \node at (6.5, 8.5) {$\varitheta$};
    \end{tikzpicture}
    \caption{$\varipi$ update}
  \end{subfigure}
  \; \; \; \;
  \begin{subfigure}[t]{0.22\textwidth}
    \begin{tikzpicture}[scale=0.28]
      \fill[white!40!white] (0,0) rectangle (3,6);
      \fill[green!90!black] (0,1) rectangle (0.5,2);
      \fill[green!90!black] (1,1) rectangle (1.5,2);
      \draw[black] (0,0) rectangle (3,6);
      \node at (1.5, 3) {$\variz$};
      \fill[white!40!white] (4,0) rectangle (9,6);
      \fill[green!90!black] (4,1) rectangle (9,2);
      \draw[black] (4,0) rectangle (9,6);
      \node at (6.5, 3) {$x$};
      \fill[white!40!white] (0,7) rectangle (3,8);
      \draw[black] (0,7) rectangle (3,8);
      \node at (-0.5, 7.5) {$\varipi$};
      \fill[white!40!white] (4,7) rectangle (9,10);
      \fill[red!40!white] (4,8.5) rectangle (9,9);
      \fill[red!40!white] (4,9.5) rectangle (9,10);
      \draw[black] (4,7) rectangle (9,10);
      \node at (6.5, 8.5) {$\varitheta$};
    \end{tikzpicture}
    \caption{$\varitheta$ update}
  \end{subfigure}
  \; \; \; \;
   \begin{subfigure}[t]{0.22\textwidth}
    \begin{tikzpicture}[scale=0.28]
      \fill[white!40!white] (0,0) rectangle (3,6);
      \fill[red!40!white] (0,1) rectangle (0.5,2);
      \fill[red!40!white] (1,1) rectangle (1.5,2);
      \draw[black] (0,0) rectangle (3,6);
      \node at (1.5, 3) {$\variz$};
      \fill[white!40!white] (4,0) rectangle (9,6);
      \fill[green!90!black] (4,1) rectangle (9,2);
      \draw[black] (4,0) rectangle (9,6);
      \node at (6.5, 3) {$x$};
      \fill[green!90!black] (0,7) rectangle (0.5,8);
      \fill[green!90!black] (1,7) rectangle (1.5,8);
      \draw[black] (0,7) rectangle (3,8);
      \node at (-0.5, 7.5) {$\varipi$};
      \fill[white!40!white] (4,7) rectangle (9,10);
      \fill[green!90!black] (4,8.5) rectangle (9,9);
      \fill[green!90!black] (4,9.5) rectangle (9,10);
      \draw[black] (4,7) rectangle (9,10);
      \node at (6.5, 8.5) {$\varitheta$};
    \end{tikzpicture}
    \caption{$\variz$ update}
  \end{subfigure}
  \vspace{1em}
  \caption{Access pattern during ESVI updates. Green indicates the
    variable or data point being read, while Red indicates it being updated.}
  \label{fig:accesspattern_esvi}
\end{figure}

In contrast, the following access pattern of ESVI allows multiple processors to access and update mutually exclusive subsets of coordinates $\Kcal$ independently (See Figure \ref{fig:accesspattern_esvi} for an illustration): 
\begin{itemize*}
\item The update for $\varipi$ \eqref{eq:varipi_update} requires
  access to the coordinates $\variz_{i, k}$ for $k \in \Kcal$.
\item The update for $\varitheta$ \eqref{eq:varin_update} and
  \eqref{eq:varinu_update} requires access to $\variz_{i,k}$ for $k \in
  \Kcal$.
\item The update to $\variz_{i,k}$ for $k \in \Kcal$ requires access to
  $\varipi_{k}$ and $\varitheta_{k}$ for $k \in \Kcal$.
\end{itemize*}

\subsection{Parallelization}
\label{sec:Parallelization}
\begin{figure*}[ht]
  \begin{subfigure}[t]{0.25\textwidth}
    \centering
    \begin{tikzpicture}[scale=0.14]

    \boxlayout
    \drawLable
    \drawXpart

    \drawThetapart{0}{3}{\ptrna}{\ptcla}
    \drawThetapart{4}{7}{\ptrnb}{\ptclb}
    \drawThetapart{8}{11}{\ptrnc}{\ptclc}
    \drawThetapart{12}{15}{\ptrnd}{\ptcld}

    \asnbox{0}{0}{red} \asnbox{1}{0}{red} \asnbox{2}{0}{red} \asnbox{3}{0}{red}
    \asnbox{4}{1}{green} \asnbox{5}{1}{green} \asnbox{6}{1}{green} \asnbox{7}{1}{green}
    \asnbox{8}{2}{blue} \asnbox{9}{2}{blue} \asnbox{10}{2}{blue} \asnbox{11}{2}{blue}
    \asnbox{12}{3}{brown} \asnbox{13}{3}{brown} \asnbox{14}{3}{brown} \asnbox{15}{3}{brown}
    
)
    
  \end{tikzpicture}
    \caption{Initial assignment of $\varitheta$ and $x$. We plot diagonal initialization while in real case random initialization is used. }
  \end{subfigure}
  \;
  \begin{subfigure}[t]{0.23\textwidth}
    \centering
    \begin{tikzpicture}[scale=0.14]

      \boxlayout
      \drawLable
      \drawXpart

      \drawThetapart{0}{0}{\ptrna}{\ptcla}
      \drawThetapart{2}{2}{\ptrna}{\ptcla}
      \drawThetapart{4}{7}{\ptrnb}{\ptclb}
      \drawThetapart{8}{11}{\ptrnc}{\ptclc}
      \drawThetapart{12}{15}{\ptrnd}{\ptcld}

      \asnbox{0}{0}{red} \asnbox{2}{0}{red}
      \asnbox{4}{1}{green} \asnbox{5}{1}{green} \asnbox{6}{1}{green} \asnbox{7}{1}{green}
      \asnbox{8}{2}{blue} \asnbox{9}{2}{blue} \asnbox{10}{2}{blue} \asnbox{11}{2}{blue}
      \asnbox{12}{3}{brown} \asnbox{13}{3}{brown} \asnbox{14}{3}{brown} \asnbox{15}{3}{brown}

      \dasnbox{1}{0} \dasnbox{1}{3} \mvarr{1}{0}{1}{3}
      \dasnbox{3}{0} \dasnbox{3}{2} \mvarr{3}{0}{3}{2}

    \end{tikzpicture}
    \caption{Worker $1$ finishes processing $\{2,4\}\in\Kcal_{1}$, it sends them over to a random worker.
      Here, $\varitheta_{2}$ is sent from worker $1$ to $4$ and $\varitheta_{4}$ from $1$ to $3$.  }
  \end{subfigure}
  \;
  \begin{subfigure}[t]{0.23\textwidth}
    \centering
    \begin{tikzpicture}[scale=0.14]

      \boxlayout
      \drawLable
      \drawXpart

      \drawThetapart{0}{0}{\ptrna}{\ptcla}
      \drawThetapart{1}{1}{\ptrnd}{\ptcld}
      \drawThetapart{2}{2}{\ptrna}{\ptcla}
      \drawThetapart{3}{3}{\ptrnc}{\ptclc}
      \drawThetapart{4}{7}{\ptrnb}{\ptclb}
      \drawThetapart{8}{11}{\ptrnc}{\ptclc}
      \drawThetapart{12}{15}{\ptrnd}{\ptcld}

      \asnbox{0}{0}{red} \asnbox{2}{0}{red}
      \asnbox{4}{1}{green} \asnbox{5}{1}{green} \asnbox{6}{1}{green} \asnbox{7}{1}{green}
      \asnbox{8}{2}{blue} \asnbox{9}{2}{blue} \asnbox{10}{2}{blue} \asnbox{11}{2}{blue} \asnbox{3}{2}{blue}
      \asnbox{12}{3}{brown} \asnbox{13}{3}{brown} \asnbox{14}{3}{brown} \asnbox{15}{3}{brown}
      \asnbox{1}{3}{brown}

    \end{tikzpicture}
    \caption{Upon receipt, the column is processed by the new worker.
      Here, worker $4$ can now operate on $\varitheta_{2}$ and $3$ on $\varitheta_{4}$}
  \end{subfigure}
  \;
  \begin{subfigure}[t]{0.23\textwidth}
    \centering
    \begin{tikzpicture}[scale=0.14]

      \boxlayout
      \drawLable
      \drawXpart

      \asnbox{0}{2}{blue} \asnbox{1}{2}{blue} \asnbox{2}{0}{red} \asnbox{3}{1}{green}
      \asnbox{4}{1}{green} \asnbox{5}{3}{brown} \asnbox{6}{0}{red} \asnbox{7}{3}{brown}
      \asnbox{8}{3}{brown} \asnbox{9}{3}{brown} \asnbox{10}{1}{green} \asnbox{11}{0}{red}
      \asnbox{12}{1}{green} \asnbox{13}{2}{blue} \asnbox{14}{0}{red} \asnbox{15}{1}{green}

      \drawThetapart{0}{1}{\ptrnc}{\ptclc}
      \drawThetapart{2}{2}{\ptrna}{\ptcla}
      \drawThetapart{3}{4}{\ptrnb}{\ptclb}
      \drawThetapart{5}{5}{\ptrnd}{\ptcld}
      \drawThetapart{6}{6}{\ptrna}{\ptcla}
      \drawThetapart{7}{9}{\ptrnd}{\ptcld}
      \drawThetapart{10}{10}{\ptrnb}{\ptclb}
      \drawThetapart{11}{11}{\ptrna}{\ptcla}
      \drawThetapart{12}{12}{\ptrnb}{\ptclb}
      \drawThetapart{13}{13}{\ptrnc}{\ptclc}
      \drawThetapart{14}{14}{\ptrna}{\ptcla}
      \drawThetapart{15}{15}{\ptrnb}{\ptclb}

    \end{tikzpicture}
    \caption{During the execution of the algorithm, the ownership of the
      global parameters $\varitheta_{k}$ changes.}
  \end{subfigure}
  \caption{Illustration of the communication pattern in asynchronous ESVI 
  algorithm (based on the NOMAD algorithm \citep{YunYuHsiVisDhi14}). Parameters of same color are in memory of the same worker. Horizontal and Vertical lines indicate the two directions of 
  partitioning data and parameters. For instance, data $x$ is partitioned horizontally along $N$ and vertically along $D$. Local parameter $\variz$ is partitioned horizontally along $N$ and vertically along $K$.
  Global parameters - $\varipi$ is partitioned vertically along $K$, and finally $\varitheta$ is partitioned horizontally along $K$ and vertically along $D$ (Here, N, D and K denote the number of data points, dimensions and  mixture components respectively).}
  \label{fig:nomad_scheme}
\end{figure*}
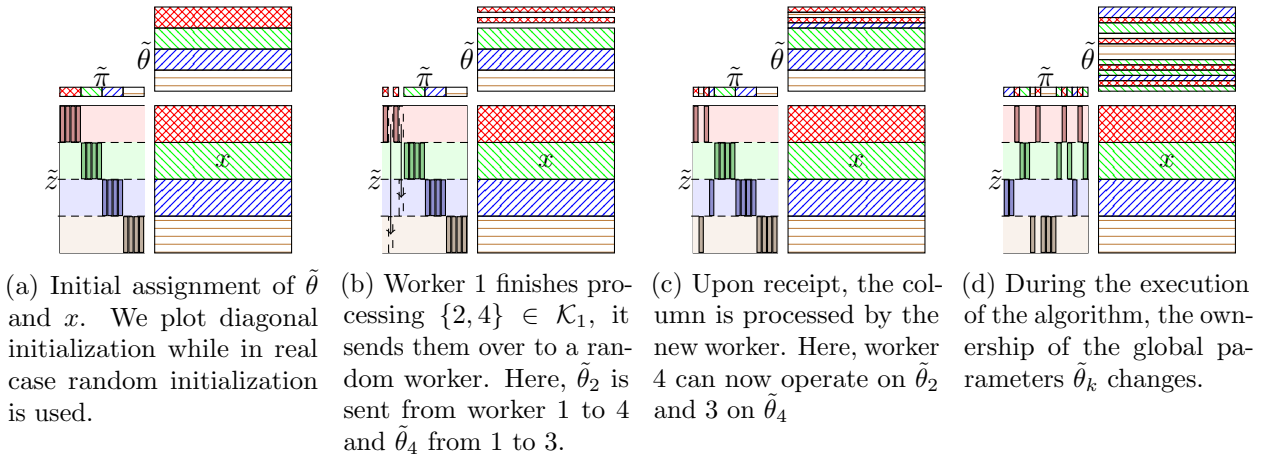

In this sub-section, we describe the parallel asynchronous algorithm of ESVI.
Let $P$ denote the number of processors, and let
$\Ical_{p} \subset \cbr{1, \ldots, N}$ denote indices of the data
points owned by processor $p$. $\variz_{i}$
for $i \in \Ical_{p}$ are local variables assigned to processor $p$. The global
variables are split across the processors. Let
$\Kcal_{p} \subset \cbr{1, \ldots, K}$ denote the indices of the rows
of $\varitheta$ currently residing in processor $p$. Then
processor $p$ can update any $\variz_{i,k}$ for $i \in \Ical_{p}$ and
$k \in \Kcal_{p}$. Finally, we need to address the issue of 
how to communicate $\varitheta_{k}$ across processors. For this, 
we follow the asynchronous communication scheme outlined by 
\citep{YunYuHsiVisDhi14} and \citep{YuHsiYunVisetal15}. Figure \ref{fig:nomad_scheme} is an illustration of how this works pictorially. We partition the data and the corresponding $\variz_{1:N}$ variables
across the processors. Each processors maintains its own queue. Once partitioned, the $\variz$ variables never move. 
On the other hand, the $\varitheta$ variables move nomadically between
processors. Each processor performs ESVI updates using the current
subset of $\varitheta$ variables that it currently holds. Then the variables
are passed on to the queue of another randomly chosen processor as shown in the second sub-figure in Figure \ref{fig:nomad_scheme}. It is this nomadic movement \citep{YunYuHsiVisDhi14} of the $\varitheta$
variables that ensures proper mixing and convergence. The complete algorithm for parallel-ESVI is outlined in Algorithm \ref{alg:parallel_esvi}.

\subsection{Comparison and Complexity}
We want to point out that ESVI updates are stochastic w.r.t. the coordinates, 
however the update in each coordinate is exact using (\ref{eq:optimal_z}). In contrast, SVI stochastically 
samples the data and performs inexact or noisy updates and does not guarantee 
each step to be an ascent step. Moreover, given a $N\times D$ dataset and 
fixing $K$ clusters, by simple calculation we can see that to update all 
$\variz_{ik}$ once, VI requires $O(DK)$ updates on $\varitheta$, while SVI 
needs $O(NDK)$ and parallel ESVI 
need $O(PDK)$. 

\begin{algorithm}[tb]
   \caption{Parallel-ESVI Algorithm}
   \label{alg:parallel_esvi}
\begin{algorithmic}
   \STATE $P$: total number of workers, \; $T$: maximum computing time
   \STATE $\Ical_{p}$: data points owned by worker $p$, \; 
   \STATE $\Kcal_{p}$: global parameters owned by worker $p$ (concurrent queue)
   \STATE Initialize global parameters $\varitheta^{0}$, $\varipi^{0}$
   \FOR{ worker $p = 1 \ldots P$ asynchronously}
   	\WHILE{Stop criteria not satisfied}
    \STATE Pick a subset $\textbf{k}_{s}\subset\Kcal_{p}$
    \FOR {All data point $i\in\Ical_{p}$}
    \FOR{ $k \in  \textbf{k}_{s}$}
		\STATE Compute $\variz_{ik}^*$ using \eqref{eq:optimal_z} 
		\STATE $\varipi_{k} += \variz^*_{ik}-\variz_{ik}$
    \STATE $\varin_{k}+=\variz^*_{ik}-\variz_{ik}$
    \STATE $\varinu_k += \rbr{\variz^*_{ik}-\variz_{ik}}\times\phi\rbr{x_i,k}$
    \STATE $\variz_{ik}\leftarrow\variz^*_{ik}$
		\STATE Pick a random worker $p'$ and send $\varipi_{k}$ and 
    $\varitheta_{k}$, push $k$ to $\Kcal_{p'}$
	\ENDFOR
  \ENDFOR
	\ENDWHILE
   \ENDFOR
\end{algorithmic}
\end{algorithm}

\section{ESVI-LDA}
\label{sec:esvilda}
In this section, we show how to apply ESVI to Latent Dirichlet Allocation (LDA). 
Recall the standard LDA model by Blei et al.\citep{BleNgJor03}. Each topic $\beta_k, k\in[K]$ is a 
distribution over the vocabulary with size $V$ and each document is a 
combination of $K$ topics. The generative process is:
\begin{itemize*}
  \item Draw topic weights $\beta_k$ $\sim$ Dirichlet(\textbf{$\eta$}), $k = 1 \ldots K$\\
  \item For every document $d_i \in \{d_1,d_2\ldots d_D\}$:
  \begin{itemize*}
    \item Draw $\theta_i$ $\sim$ Dirichlet($\alpha$)
     \item For each word $n\in[N]$:
     \begin{itemize}
      	\item Draw topic assignment $z_{in} \sim \text{Multi}(\theta_i)$
      	\item Draw word $w_{in} \sim \text{Multi}(\beta_{z_{in}})$ 
     \end{itemize}
  \end{itemize*}
\end{itemize*}
where $\alpha\in\mathbb{R}^K$ and $\eta\in\mathbb{R}^V$ are symmetric Dirichlet priors.
The inference task for LDA is to characterize the posterior distribution $p(\beta,\theta,z|w)$. While the 
posterior is intractable to compute, many methods have been developed to 
approximate the posterior. Here we use the idea in previous sections to 
develop extreme stochastic variational inference for LDA.

We denote the assignment of word $n$ in document $d_i$ as $z_{in}$ where $z_{i} \in \mathbb{R}^{K}$. 
Also $w_{in}$ denotes the $n$-th word in $i$-th document. 
Thus in LDA, the local hidden variables for a word is the word assignment vector $z_{in}$ and local hidden variable for a document is $z_i$ and the topic mixture $\theta_i$. The global hidden variable are the topics $\beta_k$.
Given these, we can formulate the complete conditional of the topics $\beta_k$ $\theta_i$ and $z_{in}$ as:
\begin{align*}
p(\beta_k|z,w)  & =  \text{Dirichlet}(\eta+\sum_{i=1}^{D}\sum_{n=1}^{N} z_{in}^k w_{in}), \\
p(\theta_i|z_i)  & = \text{Dirichlet}(\alpha + \sum_{n=1}^N z_{in}) \\
p(z_{in}^k=1 | \theta_i, \beta_{1:K}, w_{in}) & \propto \exp\rbr{\log\theta_{ik} + \log\beta_k^{w_{in}}}
\end{align*}
We denote multinomial parameter for $z_{in}^k$ as $\phi_{in}^k$, 
Dirichlet parameter for $\beta_k$ and $\theta_i$ as $\lambda_k$ and 
$\gamma_i$. The update rules for these three variational parameters are:\\
\begin{align*}
& \lambda_k = \eta+\sum_{i=1}^{D}\sum_{n=1}^{N} z_{in}^k w_{in},
\quad \gamma_i = \alpha + \sum_{n=1}^N z_{in}\\
& \phi_{in}^k \propto \exp\rbr{ \Psi(\gamma_i^k) + \Psi(\lambda_k^{w_{in}}) - 
\Psi(\sum_{v=1}^V \lambda_k^v)}
\end{align*}

where $\Psi$ is the digamma function and we denote $\pi_k=\sum_{v=1}^V \lambda_k^v$. Traditional VI algorithms infer all the local variables $\theta$, $z$ and then update the global variable $\beta$. This 
is very inefficient and not scalable. Notice that when updating 
$\phi_{in}^k$ we only need to access $\gamma_i^k$, $\lambda_k^{w_{in}}$ and $\pi_k$. 
And similarly, once $\phi_{in}^k$ is modified, the parameters that need to be updated are 
$\gamma_i^k$, $\lambda_k^{w_{in}}$ and $\pi_k$. Therefore, as long as $\pi_k$ 
can be accessed, the updates to these parameters can be parallelized. 
Based on the ideas we introduced in Section \ref{sec:ExtrStochVari}, 
we propose an asynchronous distributed method ESVI-LDA, which is outlined in Algorithm \ref{alg:esvilda}. 
Besides working threads, each machine also has a sender thread and a receiver 
thread, which enables the non-locking send/recv of parameters.
One key issue here is how to keep $\pi_{1:K}$ up-to-date across multiple 
processors. For this, we follow \citep{YuHsiYunVisetal15}, who present a 
scheme for keeping a slowly changing $K$ dimensional vector, approximately 
synchronized across multiple machines. Succinctly, the idea is to 
communicate the changes in $\pi$ using a round robin fashion. Since 
$\pi$ does not change rapidly, one can tolerate some staleness without 
adversely affecting convergence. 

\begin{algorithm}[th]
\caption{ESVI-LDA Algorithm}
\label{alg:esvilda}
\begin{algorithmic}
   \STATE Load $\{d_1\ldots d_D\}$ into $P$ machines 
   \STATE Initialize $\phi$, $\gamma$, $\lambda$ using priors $\alpha, \eta$
   \STATE Initialize job queue $Q$: distribute $\lambda^{1:V}$ in $P$ machines
   \STATE Initialize sender queue $q_s$
   \FOR{every machine asynchronously}
    \IF{receiver thread}
      \WHILE{receive $\lambda^v$} 
        \STATE $push\rbr{Q_t,\lambda^v}$ for some $t$
      \ENDWHILE
    \ENDIF
    \IF{sender thread}
      \WHILE{not $q_s.empty\rbr{}$}
        \STATE send $q_s.pop()$ to next random machine
      \ENDWHILE
    \ENDIF
    \IF{worker thread $t$}
    \STATE pop from $Q_t$: $\lambda^v$, 
    \FOR{all local word token s.t. $w_{dn}=v$}
      \FOR{ $k=1 \ldots K$}
        \STATE $\phi_{dn}^{k} \propto {\exp \rbr{\psi\rbr{\gamma_d^k} + \psi\rbr{\lambda_k^{w_{dn}}} - \psi\rbr{\sum_{v} \lambda_k^v} }}$
      \ENDFOR
      \FOR{ $k=1 \ldots K$}
        \STATE $\gamma_{d}^{k} += \phi_{dn}^k - \phi_{dn}^k(old)$ 
        \STATE $\lambda_k^{w_{dn}} += \phi_{dn}^k - \phi_{dn}^k(old)$
      \ENDFOR
		  \STATE Update global $\sum_{v} \lambda_k^v$
    \ENDFOR
    \STATE $q_s.push\rbr{\lambda^v}$
    \ENDIF
   \ENDFOR
\end{algorithmic}
\end{algorithm}

In order to update $\phi_{in}^k$ we need only to
access $\gamma_i^k$ $\lambda_k^{w_{in}}$ and $\pi_k$. And similarly, 
once $\phi_{in}^k$ is modified, only parameters  
$\gamma_i^k$, $\lambda_k^{w_{in}}$ and $\pi_k$ need to be updated. 
Following that, for each word token, 
these parameters can be updated independently. In our setting, each machine loads its
own chunk of the data, and also has local model parameters $\gamma$ 
and $\phi$. Each machine maintains a local job queue that stores global 
parameters $\lambda$ that is now owned by this machine. After updating with 
each $\lambda_{1:K}^v$, the machine sends it to another machine while pushing 
$v$ into the job queue of that machine. This leads to a fully asynchronous and 
non-locking distributed algorithm.

{\bf Handling large number of topics:} In VI for LDA, the linear dependence of the 
model size on $K$ prevents scaling to large $K$ due to memory limitations. 
Our ESVI-LDA-TOPK approach addresses this: instead
of storing all $K$ components of the assignment parameter, we only store the 
most important top $k$ topics (using a min-heap of size $k$).
By only maintaining $C\ll K$ top values, we get performance very close to 
storing all the values staying within the memory limit. 


\section{Experiments}
\label{sec:expmts}
In our experiments, we compare our proposed {\bf ESVI-GMM} and {\bf ESVI-LDA} 
methods against VI and SVI. To handle large number of topics in LDA, we also implemented a more efficient version {\bf ESVI-LDA-TOPK}. Details are in Appendix \ref{sec:esvilda}. We use real-world datasets of varying scale as described in Table \ref{table:datasets}.
We used a large-scale parallel computing platform with
node configuration of 20 Intel Xeon E5-2680 CPUs and 256 GB memory. We implemented ESVI-LDA in C++ using MPICH, OpenMP and Intel TBB. For Distributed-VI and SVI 
implementations, we modified the authors' original code in C\footnote{\url{http://www.cs.princeton.edu/~blei/lda-c/}. Distributed-VI was implemented in Map-Reduce style}.

\begin{table}[htp]
\centering
  \renewcommand{\arraystretch}{1}
  \scalebox{0.9}{
\begin{tabular}{c|c|c|c} 
 & \# documents & \# vocabulary & \#words \\ \hline 
AP-DATA & 2,246 & 10,473 & 912,732 \\ 
NIPS & 1,312 & 12,149 & 1,658,309 \\ 
Enron & 37,861 & 28,102 & 6,238,796 \\ 
Ny Times & 298,000 & 102,660 & 98,793,316 \\ 
PubMed & 8,200,000 & 141,043 & 737,869,083 \\ 
UMBC-3B & 40,599,164 & 3,431,260 & 3,013,004,127 \\
\end{tabular}}
\caption{Data Characteristics}
\label{table:datasets}
\end{table}

\subsection{ESVI-GMM}
\label{sec:gmm_experiments}
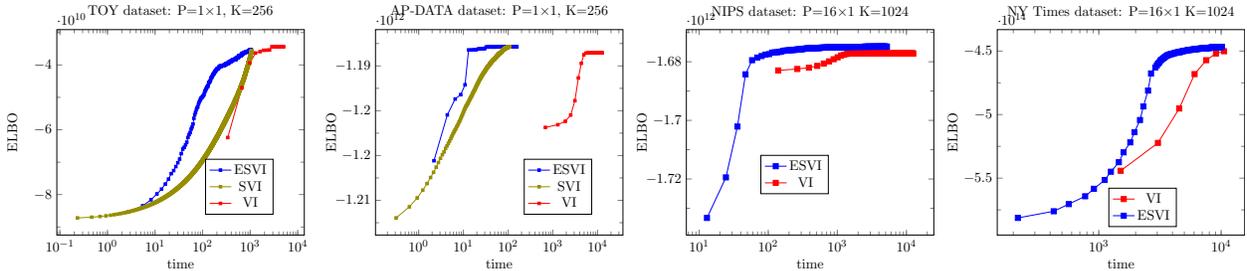
\begin{figure}[H]
        \begin{tikzpicture}[scale=0.48]
        \begin{axis}[xmode=log, minor tick num=1,
          legend style={at={(0.6,0.1)}, anchor= south west},
          title={TOY dataset: P=1$\times$1, K=256},
          xlabel={time}, ylabel={ELBO}]
	
            \addplot[thin, mark=square*, mark size=1pt, color=blue] table [x index=1, y index=3,
            header=true, col sep = comma]
            {./figuredata/GMM/TOY/ESVI_GMM.1P1.K256.V300000.txt};
            \addlegendentry{ESVI}                        

            \addplot[thin, solid, mark=square*, mark size=1pt, color=olive] table [x index=1, y index=3,
            header=true, col sep = comma]
            {./figuredata/GMM/TOY/SVI_GMM.1P1.K256.V300000.txt};
            \addlegendentry{SVI}
            
            \addplot[thin, solid, mark=square*, mark size=1pt, color=red] table [x index=1, y index=3,
            header=true, col sep = comma]
            {./figuredata/GMM/TOY/VI_GMM.1P1.K256.V300000.txt};
            \addlegendentry{VI}
            
        \end{axis}
      \end{tikzpicture}
      \begin{tikzpicture}[scale=0.48]
        \begin{axis}[xmode=log, minor tick num=1,
          legend style={at={(0.6,0.1)}, anchor= south west},
          title={AP-DATA dataset: P=1$\times$1, K=256},
          xlabel={time}, ylabel={ELBO}]
	
            \addplot[thin, mark=square*, mark size=1pt, color=blue] table [x index=1, y index=3,
            header=true, col sep = comma]
            {./figuredata/GMM/AP_DATA/ESVI_GMM.1P1.K256.V300000.txt};
            \addlegendentry{ESVI}                        

            \addplot[thin, solid, mark=square*, mark size=1pt, color=olive] table [x index=1, y index=3,
            header=true, col sep = comma]
            {./figuredata/GMM/AP_DATA/SVI_GMM.1P1.K256.V300000.txt};
            \addlegendentry{SVI}
            
            \addplot[thin, solid, mark=square*, mark size=1pt, color=red] table [x index=1, y index=3,
            header=true, col sep = comma]
            {./figuredata/GMM/AP_DATA/VI_GMM.1P1.K256.V300000.txt};
            \addlegendentry{VI}
            
        \end{axis}
      \end{tikzpicture}        
      \begin{tikzpicture}[scale=0.48]
        \begin{axis}[xmode=log, minor tick num=1,
          legend style={at={(0.3,0.2)}, anchor= south west},
          title={NIPS dataset: P=16$\times$1 K=1024},
          xlabel={time}, ylabel={ELBO}]

            \addplot[thin, mark=square*, mark size=2pt, color=blue] table [x index=1, y index=3,
            header=true, col sep = comma]
            {./figuredata/GMM/NIPS/ESVI_GMM.16P1.K1024.V50000.txt};
            \addlegendentry{ESVI}                        

            \addplot[thin, solid, mark=square*, mark size=2pt, color=red] table [x index=1, y index=3,
            header=true, col sep = comma]
            {./figuredata/GMM/NIPS/VI.16P1.K1024.W0.5.V50000.txt};
            \addlegendentry{VI}
        \end{axis}
      \end{tikzpicture} 
      \begin{tikzpicture}[scale=0.48]
        \begin{axis}[xmode=log, minor tick num=1,
          legend style={at={(0.45,0.04)}, anchor= south west},
          title={NY Times dataset: P=16$\times$1 K=1024},
          xlabel={time}, ylabel={ELBO}]
          
            \addplot[thin, mark=square*, mark size=2pt, color=red] table [x index=1, y index=3,
            header=true, col sep = comma]
            {./figuredata/GMM/NYTIMES/VI.16P1.K1024.W0.5.V5000000.txt};
            \addlegendentry{VI}                        

            \addplot[thin, solid, mark=square*, mark size=2pt, color=blue] table [x index=1, y index=3,
            header=true, col sep = comma]
            {./figuredata/GMM/NYTIMES/ESVI_GMM.16P1.K1024.V5000000.txt};
            \addlegendentry{ESVI}
            
        \end{axis}
      \end{tikzpicture} 
      \caption{Comparison of ESVI-GMM, SVI and VI. $P=N\times n$ 
      denotes $N$ machines each with $n$ threads.}
\label{fig:multimachine_gmm}
\end{figure}

In this section, we first compare ESVI-GMM with SVI and VI in the Single Machine Single thread setting. We use a TOY dataset which consists of $N=29,983$ data points, $D=128$ dimensions and the AP-DATA dataset which consists of $N=2,246$ data points, $D=10,473$ dimensions. In both cases, we set the number components $K=256$. We plot the performance of the methods (ELBO) as a function of time. ESVI-GMM outperforms SVI and VI by quite some margin. For Multi Machine case, we use the NIPS and NY Times datasets and only compare against VI (SVI does not apply; it needs to update all its $K$ global parameters which is infeasible when $K$ is large). Although typically these datasets do not demand running on multiple machines, out intention here is to demonstrate scalability to very large number of components ($K=1024$) and dimensions, which is typically the case in large scale text datasets with millions of word count features. Traditionally, GMM inference methods have not been able to handle such a scale. Results in Figure \ref{fig:multimachine_gmm} give a clear indication that ESVI-GMM is able to outperform VI by a clear margin.

\subsection{ESVI-LDA}
\label{sec:lda_experiments}

\subsubsection{Single Machine Single thread}
\label{subsec:singlemachine_singlethread}
We compare serial versions of the methods on Enron and NY Times datasets which are medium sized
and fit on a single-machine. On both datasets, we run with single machine and single thread. For Enron, we set \# of topics $K=8, 16, 20, 32, 64, 128, 256$. For NY Times, we set $K=8, 16, 32, 64$. To keep the plots concise, we only show results with $K=16, 64, 128$ in Figure \ref{fig:singlemachine_singlethread} (two left-most plots). ESVI-LDA performs better than VI and SVI in both the datasets for all values of $K$. In our experiments, x-axis is in log-scale. 

\subsubsection{Single Machine Multi Core}
\label{subsec:singlemachine_multicore}
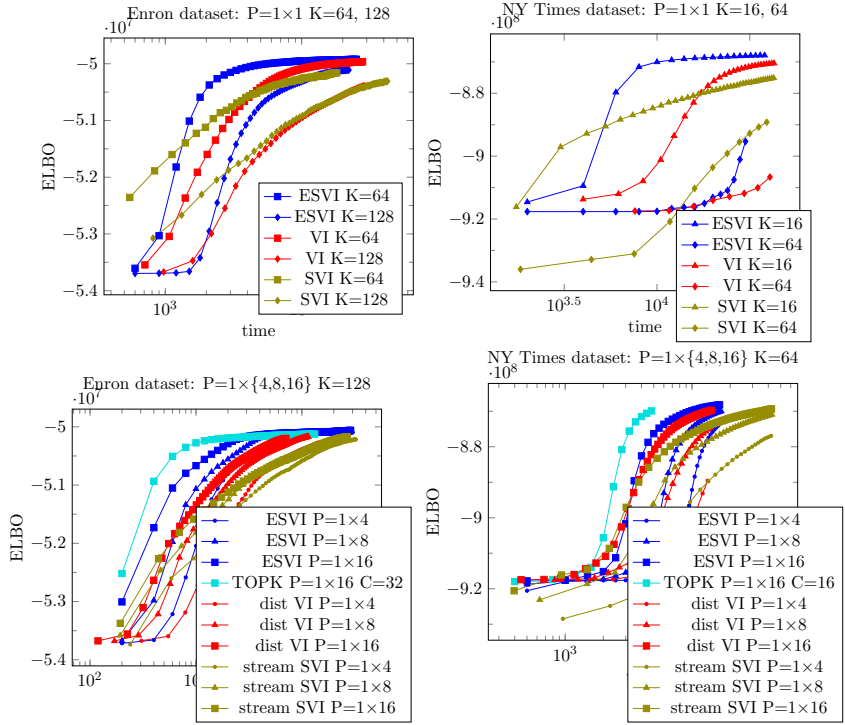
\begin{figure}[H]
 \centering
      \begin{tikzpicture}[scale=0.6]
      \begin{axis}[xmode=log, minor tick num=1,
	legend style={at={(0.5,0.18)}, anchor= west},
          title={Enron dataset: P=1$\times$1 K=64, 128},
          restrict y to domain=-6e7:-4.2e7,
          xlabel={time}, ylabel={ELBO}, name=]
          
            \addplot[thin, mark=square*, mark size=2pt, color=blue] table [skip first n=2, x index=1, y index=3,
            header=true, col sep = comma]
            {./figuredata/SINGLE_MACHINE/ENRON/ESVI.1P1.K64.txt};
            \addlegendentry{ESVI K=64}                        

            \addplot[thin, mark=diamond*, mark size=2pt, color=blue] table [skip first n=2, x index=1, y index=3,
            header=true, col sep = comma]
            {./figuredata/SINGLE_MACHINE/ENRON/ESVI.1P1.K128.txt};
            \addlegendentry{ESVI K=128}                        

%
                            
            \addplot[thin, color=red, mark=square*, mark size=2pt] table [skip first n=2, x index=1, y index=3,
            header=true, col sep = comma]
            {./figuredata/SINGLE_MACHINE/ENRON/VI_enron.1P1.K64.txt};
            \addlegendentry{VI K=64}                        

            \addplot[thin, color=red, mark=diamond*, mark size=2pt] table [skip first n=2, x index=1, y index=3,
            header=true, col sep = comma]
            {./figuredata/SINGLE_MACHINE/ENRON/VI_enron.1P1.K128.txt};
            \addlegendentry{VI K=128}
            
            \addplot[thin, color=olive, mark=square*, mark size=2pt] table [skip first n=2, x index=1, y index=3,
            header=true, col sep = comma]
            {./figuredata/SINGLE_MACHINE/ENRON/distSVI.1P1.K64_noperp_inftime.txt};
            \addlegendentry{SVI K=64}                        

            \addplot[thin, color=olive, mark=diamond*, mark size=2pt] table [skip first n=2, x index=1, y index=3,
            header=true, col sep = comma]
            {./figuredata/SINGLE_MACHINE/ENRON/distSVI.1P1.K128_noperp_inftime.txt};
            \addlegendentry{SVI K=128}
                                    
        \end{axis}
      \end{tikzpicture}
      \begin{tikzpicture}[scale=0.6]
        \begin{axis}[xmode=log, minor tick num=1,
          legend style={at={(0.6,-0.2)}, anchor= south west},
          title={NY Times dataset: P=1$\times$1 K=16, 64},
          restrict y to domain=-1.1e9:-0.7e9,
          xlabel={time}, ylabel={ELBO},name=]

            \addplot[thin, mark=triangle*, mark size=2pt, color=blue] table [x index=1, y index=3,
            header=true, col sep = comma]
            {./figuredata/SINGLE_MACHINE/NYTIMES/ESVI.1P1.K16.txt};
            \addlegendentry{ESVI K=16}
            
            \addplot[thin, mark=diamond*, mark size=2pt, color=blue] table [x index=1, y index=3,
            header=true, col sep = comma]
            {./figuredata/SINGLE_MACHINE/NYTIMES/ESVI.1P1.K64.txt};
            \addlegendentry{ESVI K=64}                        
            
            \addplot[thin, mark=triangle*, mark size=2pt, color=red] table [x index=1, y index=3,
            header=true, col sep = comma]
            {./figuredata/SINGLE_MACHINE/NYTIMES/VI.1P1.K16.txt};
            \addlegendentry{VI K=16}
            
            \addplot[thin, mark=diamond*, mark size=2pt, color=red] table [x index=1, y index=3,
            header=true, col sep = comma]
            {./figuredata/SINGLE_MACHINE/NYTIMES/VI.1P1.K64.txt};
            \addlegendentry{VI K=64}
            
            \addplot[thin, mark=triangle*, mark size=2pt, color=olive] table [x index=1, y index=3,
            header=true, col sep = comma]
            {./figuredata/SINGLE_MACHINE/NYTIMES/distSVI.1P1.K16_noperp_inftime.txt};
            \addlegendentry{SVI K=16}
            
            \addplot[thin, mark=diamond*, mark size=2pt, color=olive] table [x index=1, y index=3,
            header=true, col sep = comma]
            {./figuredata/SINGLE_MACHINE/NYTIMES/distSVI.1P1.K64_noperp_inftime.txt};
            \addlegendentry{SVI K=64}
                        
        \end{axis}
      \end{tikzpicture}
      
        \begin{tikzpicture}[scale=0.6]
        \begin{axis}[xmode=log, minor tick num=1,
          legend style={at={(0.4,-0.22)}, anchor= south west},
          title={Enron dataset: P=1$\times$\{4,8,16\} K=128},
          restrict y to domain=-6e7:-4e7,
          xlabel={time}, ylabel={ELBO}, name=]
            \addplot[thin, mark=*, mark size=1pt, color=blue] table [x index=1, y index=3,
            header=true, col sep = comma]
            {./figuredata/SINGLE_MACHINE/ENRON/ESVI.1P4.K128.txt};
            \addlegendentry{ESVI P=1$\times$4}
            
            \addplot[thin, mark=triangle*, mark size=2pt, color=blue] table [x index=1, y index=3,
            header=true, col sep = comma]
            {./figuredata/SINGLE_MACHINE/ENRON/ESVI.1P8.K128.txt};
            \addlegendentry{ESVI P=1$\times$8}
            
            \addplot[thin, mark=square*, mark size=2pt, color=blue] table [x index=1, y index=3,
            header=true, col sep = comma]
            {./figuredata/SINGLE_MACHINE/ENRON/ESVI.1P16.K128.txt};
            \addlegendentry{ESVI P=1$\times$16}     
            
            \addplot[thin, mark=square*, mark size=2pt, color=blue-green] table [x index=1, y index=3,
            header=true, col sep = comma]
	    {./figuredata/SINGLE_MACHINE/ENRON/TOPK.1P16.K128.C32.txt};
            \addlegendentry{TOPK P=1$\times$16 C=32}                        

            \addplot[thin, mark=*, mark size=1pt, color=red] table [x index=1, y index=3,
            header=true, col sep = comma] 
            {./figuredata/SINGLE_MACHINE/ENRON/VI.1P4.K128.docpara.txt};
            \addlegendentry{dist VI P=1$\times$4}
            
            \addplot[thin, mark=triangle*, mark size=2pt, color=red] table [x index=1, y index=3,
            header=true, col sep = comma]
            {./figuredata/SINGLE_MACHINE/ENRON/VI.1P8.K128.docpara.txt};
            \addlegendentry{dist VI P=1$\times$8}
            
            \addplot[thin, mark=square*, mark size=2pt, color=red] table [x index=1, y index=3,
            header=true, col sep = comma]
            {./figuredata/SINGLE_MACHINE/ENRON/VI.1P16.K.docpara.txt};
            \addlegendentry{dist VI P=1$\times$16}
            
            \addplot[thin, mark=*, mark size=1pt, color=olive] table [x index=1, y index=3,
            header=true, col sep = comma] 
            {./figuredata/SINGLE_MACHINE/ENRON/distSVI.1P4.K128_noperp_inftime.txt};
            \addlegendentry{stream SVI P=1$\times$4}
            
            \addplot[thin, mark=triangle*, mark size=2pt, color=olive] table [x index=1, y index=3,
            header=true, col sep = comma]
            {./figuredata/SINGLE_MACHINE/ENRON/distSVI.1P8.K128_noperp_inftime.txt};
            \addlegendentry{stream SVI P=1$\times$8}
            
            \addplot[thin, mark=square*, mark size=2pt, color=olive] table [x index=1, y index=3,
            header=true, col sep = comma]
            {./figuredata/SINGLE_MACHINE/ENRON/distSVI.1P16.K128_noperp_inftime.txt};
            \addlegendentry{stream SVI P=1$\times$16}
            
        \end{axis}
      \end{tikzpicture} 
      \begin{tikzpicture}[scale=0.6]
        \begin{axis}[xmode=log, minor tick num=1,
          legend style={at={(0.45,-0.32)}, anchor= south west},
          title={NY Times dataset: P=1$\times$\{4,8,16\} K=64},
          restrict y to domain=-1.1e9:-0.8e9,
          xlabel={time}, ylabel={ELBO},name=]
            \addplot[thin, mark=*, mark size=1pt, color=blue] table [x index=1, y index=3,
            header=true, col sep = comma]
            {./figuredata/SINGLE_MACHINE/NYTIMES/ESVI.1P4.K64.txt};
            \addlegendentry{ESVI P=1$\times$4}
            
            \addplot[thin, mark=triangle*, mark size=2pt, color=blue] table [x index=1, y index=3,
            header=true, col sep = comma]
            {./figuredata/SINGLE_MACHINE/NYTIMES/ESVI.1P8.K64.txt};
            \addlegendentry{ESVI P=1$\times$8}
            
            \addplot[thin, mark=square*, mark size=2pt, color=blue] table [x index=1, y index=3,
            header=true, col sep = comma]
            {./figuredata/SINGLE_MACHINE/NYTIMES/ESVI.1P16.K64.txt};
            \addlegendentry{ESVI P=1$\times$16}                        
            
            \addplot[thin, mark=square*, mark size=2pt, color=blue-green] table [x index=1, y index=3,
            header=true, col sep = comma]
            {./figuredata/SINGLE_MACHINE/NYTIMES/TOPK.1P16.K64.C16.txt};            
            \addlegendentry{TOPK P=1$\times$16 C=16}                        

            \addplot[thin, mark=*, mark size=1pt, color=red] table [x index=1, y index=3,
            header=true, col sep = comma]
            {./figuredata/SINGLE_MACHINE/NYTIMES/VI.1P4.K64.docpara.txt};
            \addlegendentry{dist VI P=1$\times$4}
            
            \addplot[thin, mark=triangle*, mark size=2pt, color=red] table [x index=1, y index=3,
            header=true, col sep = comma]
            {./figuredata/SINGLE_MACHINE/NYTIMES/VI.1P8.K64.docpara.txt};
            \addlegendentry{dist VI P=1$\times$8}
            
            \addplot[thin, mark=square*, mark size=2pt, color=red] table [x index=1, y index=3,
            header=true, col sep = comma]
            {./figuredata/SINGLE_MACHINE/NYTIMES/VI.1P16.K64.docpara.txt};
            \addlegendentry{dist VI P=1$\times$16}
            
            \addplot[thin, mark=*, mark size=1pt, color=olive] table [x index=1, y index=3,
            header=true, col sep = comma]
            {./figuredata/SINGLE_MACHINE/NYTIMES/distSVI.1P4.K64_noperp_inftime.txt};
            \addlegendentry{stream SVI P=1$\times$4}
            
            \addplot[thin, mark=triangle*, mark size=2pt, color=olive] table [x index=1, y index=3,
            header=true, col sep = comma]
            {./figuredata/SINGLE_MACHINE/NYTIMES/distSVI.1P8.K64_noperp_inftime.txt};
            \addlegendentry{stream SVI P=1$\times$8}
            
            \addplot[thin, mark=square*, mark size=2pt, color=olive] table [x index=1, y index=3,
            header=true, col sep = comma]
            {./figuredata/SINGLE_MACHINE/NYTIMES/distSVI.1P16.K64_noperp_inftime.txt};
            \addlegendentry{stream SVI P=1$\times$16}            
            
        \end{axis}
      \end{tikzpicture}        	 
\caption{Single Machine experiments for ESVI-LDA (Single and Multi core). TOPK refers to our ESVI-TOPK method. $P=N\times n$ denotes $N$ machines each with $n$ threads.}

  \label{fig:singlemachine_singlethread}
\end{figure}

We evaluate the performance of distributed ESVI-LDA against a map-reduce based distributed implementation of VI, and the streaming SVI method \citep{BroBoyWibWilJor13}. We vary the number of cores as $4, 8, 16$. This is shown in Figure \ref{fig:singlemachine_singlethread} (two right-most plots). For Enron dataset, we use $K=128$ and for NY Times dataset, we use $K=64$. ESVI-LDA outperforms VI and SVI consistently in all scenarios. In addition, we observe that both the methods benefit reasonably when we provide more cores to the computation. We observe that ESVI-LDA-TOPK, which stores only top 1/4-th of $K$ topics performs the best on both datasets. 

\subsubsection{Multi Machine Multi Core}
\label{subsec:multimachine_multicore}
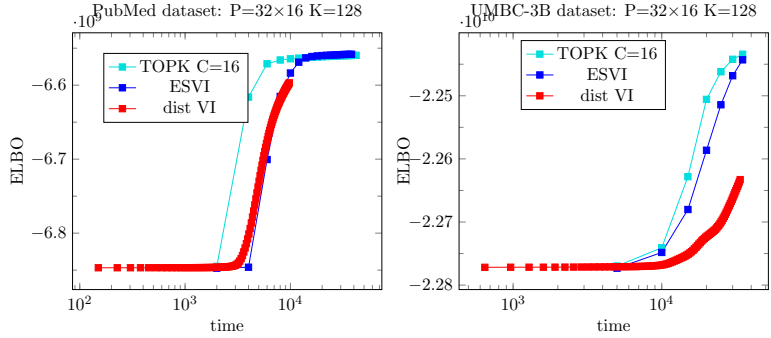
\begin{figure}[H]
 \centering
      \begin{tikzpicture}[scale=0.6]
        \begin{axis}[xmode=log, minor tick num=1,
          legend style={at={(0.1,0.65)}, anchor= south west},
          title={PubMed dataset: P=32$\times$16 K=128},
          restrict y to domain=-8e9:5,
          xlabel={time}, ylabel={ELBO}]

            \addplot[thin, mark=square*, mark size=2pt, color=blue-green] table [x index=1, y index=3,
            header=true, col sep = comma]
            {./figuredata/MULTI_MACHINE/PUBMED/TOPK.32P16.K128.C16.hard.txt};
            \addlegendentry{TOPK C=16}

            \addplot[thin, mark=square*, mark size=2pt, color=blue] table [x index=1, y index=3,
            header=true, col sep = comma]
            {./figuredata/MULTI_MACHINE/PUBMED/ESVI.32P16.K128.hard.txt};            
            \addlegendentry{ESVI}

            \addplot[thin, mark=square*, mark size=2pt, color=red] table [x index=1, y index=3,
            header=true, col sep = comma]
            {./figuredata/MULTI_MACHINE/PUBMED/VI.32P16.K128.docpara.txt};
            \addlegendentry{dist VI}            
            
        \end{axis}
      \end{tikzpicture} 
      \begin{tikzpicture}[scale=0.6]
        \begin{axis}[xmode=log, minor tick num=1,
          legend style={at={(0.2,0.7)}, anchor= south west},
          title={UMBC-3B dataset: P=32$\times$16 K=128},
          restrict y to domain=-4.5e10:5,
          xlabel={time}, ylabel={ELBO}]

            \addplot[thin, solid, mark=square*, mark size=2pt, color=blue-green] table [x index=1, y index=3,
            header=true, col sep = comma]
            {./figuredata/MULTI_MACHINE/UMBC/TOPK.32P16.K128.C16.hard.txt};
            \addlegendentry{TOPK C=16}
            
            \addplot[thin, mark=square*, mark size=2pt, color=blue] table [x index=1, y index=3,
            header=true, col sep = comma]
            {./figuredata/MULTI_MACHINE/UMBC/ESVI.32P16.K128.hard.txt};
            \addlegendentry{ESVI}                        

            \addplot[thin, mark=square*, mark size=2pt, color=red] table [x index=1, y index=3,
            header=true, col sep = comma]
            {./figuredata/MULTI_MACHINE/UMBC/VI.32P16.K128.docpara.txt};
            \addlegendentry{dist VI}            
            
        \end{axis}
      \end{tikzpicture} 
\caption{Multi Machine Multi Core experiments for ESVI-LDA. TOPK is our ESVI-TOPK method}
\label{fig:multimachine_multicore}
\end{figure}

We stretch the limits of ESVI-LDA method and compare it against distributed VI on 
large datasets: PubMed and UMBC-3B. UMBC-3B is a massive dataset with 3 billion tokens and a vocabulary of 3 million. Here, we make use of 32 nodes and 16 cores and learn number of topics $K=128$. As the results in Figure \ref{fig:multimachine_multicore} demonstrate, ESVI-LDA achieves a better solution than distributed VI in all cases. On the largest dataset UMBC-3B, ESVI-LDA is also much faster than VI. In PubMed, VI has a slight initial advantage, however eventually ESVI progresses much faster towards a better ELBO. ESVI-LDA-TOPK is particularly better than the other two on both the datasets. On PubMed especially, the top-k approach gives us significant gains in time. 

\subsubsection{Predictive Performance}
\label{subsec:pred_performance}
\begin{figure}[H]
\centering
       \begin{tikzpicture}[scale=0.65]
        \begin{axis}[xmode=log, minor tick num=1,
          title={Enron dataset: P=1$\times$16 K=32},
          xlabel={time}, ylabel={Test Perplexity}]

            \addplot[thin, mark=square*, mark size=2pt, color=blue] table [x index=1, y index=4,
            header=true, col sep = comma]
            {./figuredata/SINGLE_MACHINE/ENRON_TEST/ESVI.1P16.K32.S1024.txt};            
            \addlegendentry{ESVI}

            \addplot[thin, mark=square*, mark size=2pt, color=red] table [x index=1, y index=4,
            header=true, col sep = comma]
            {./figuredata/SINGLE_MACHINE/ENRON_TEST/VI.1P16.K32.S1024.txt};
            \addlegendentry{VI}            
            
        \end{axis}
      \end{tikzpicture} 
      \begin{tikzpicture}[scale=0.65]
        \begin{axis}[xmode=log, minor tick num=1,
          title={NY Times dataset: P=1$\times$16 K=32},
          xlabel={time}, ylabel={Test Perplexity}]

            \addplot[thin, mark=square*, mark size=2pt, color=blue] table [x index=1, y index=4,
            header=true, col sep = comma]
            {./figuredata/SINGLE_MACHINE/NYTIMES_TEST/ESVI.1P16.K32.S16384.txt};
            \addlegendentry{ESVI}                        

            \addplot[thin, mark=square*, mark size=2pt, color=red] table [x index=1, y index=4,
            header=true, col sep = comma]
            {./figuredata/SINGLE_MACHINE/NYTIMES_TEST/VI.16P1.K32.S16384.txt};
            \addlegendentry{VI}            
            
        \end{axis}
      \end{tikzpicture} 
\caption{Predictive Performance of ESVI-LDA}
\label{fig:predictive_performance}
\end{figure}
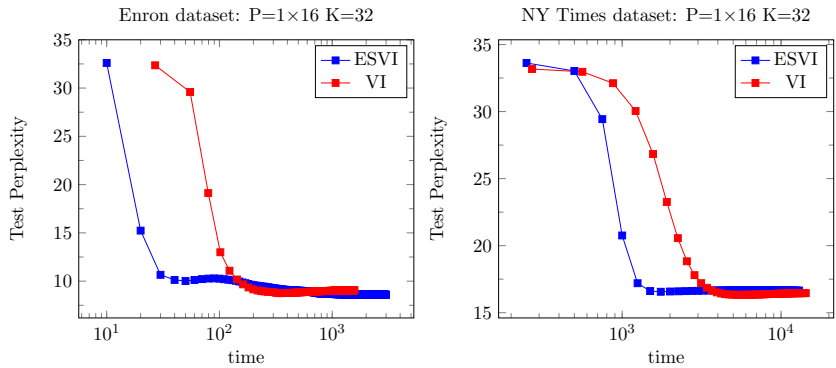

We evaluate the predictive performance of ESVI-LDA comparing against distributed VI on Enron and NY Times datasets on multiple cores. As shown in Figure \ref{fig:predictive_performance}, ESVI typically reaches comparable test perplexity scores as VI but in much shorter wallclock time. 

\section{Conclusion}
\label{sec:Conclusion}
In this paper, we proposed Extreme Stochastic Variational Inference (ESVI), a distributed, 
asynchronous and lock-free algorithm to perform large-scale inference for mixture of 
exponential families. ESVI exhibits both model as well as data parallelism simultaneously, 
allowing us to handle real-world datasets with large number of documents as well as learn 
sufficiently large number of parameters. To demonstrate its practical applicability, we show how to use 
ESVI to fit GMM and LDA models on large scale real-world datasets consisting of millions of terms and billions of documents. Our extensive empirical study is strongly suggestive that 
ESVI outperforms VI and SVI, and in most cases achieves a better quality 
solution. As future work, ESVI framework can be extended to several other latent variables models such as Stochastic Block Models and Bayesian Matrix Factorization. 

\bibliographystyle{abbrv}
\bibliography{esvi,bibfile}

\clearpage 
\newpage
\appendix


\section{Effect of using the top-k strategy}
\label{sec:varying_C_topk}

\subsubsection{Effect of varying $C$ (cutoff for k) in ESVI-LDA-TOPK}
\label{subsec:topk}
\begin{figure}[ht]
 \centering
      \begin{tikzpicture}[scale=0.6]
        \begin{axis}[xmode=log, minor tick num=1,
          legend style={at={(0.4,0.37)}, anchor= south west},
          title={Enron dataset: P=1$\times$16 K=128 C=\{1, 8, 32, 64, 128\}},
          xlabel={time}, ylabel={ELBO}]
            \addplot[thick, mark=*, mark size=1pt, color=red] table [x index=1, y index=3,
            header=true, col sep = comma]
            {./figuredata/SINGLE_MACHINE/ENRON/TOPK.1P16.K128.C1.txt};
            \addlegendentry{ESVI-TOPK C=1}
            
            \addplot[thick, mark=triangle*, mark size=1pt, color=green] table [x index=1, y index=3,
            header=true, col sep = comma]
            {./figuredata/SINGLE_MACHINE/ENRON/TOPK.1P16.K128.C8.txt};
            \addlegendentry{ESVI-TOPK C=8}
            
            \addplot[thick, mark=square*, mark size=1pt, color=brown] table [x index=1, y index=3,
            header=true, col sep = comma]
            {./figuredata/SINGLE_MACHINE/ENRON/TOPK.1P16.K128.C32.txt};
            \addlegendentry{ESVI-TOPK C=32}
            
            \addplot[thick, mark=diamond*, mark size=1pt, color=magenta] table [x index=1, y index=3,
            header=true, col sep = comma]
            {./figuredata/SINGLE_MACHINE/ENRON/TOPK.1P16.K128.C64.txt};
            \addlegendentry{ESVI-TOPK C=64}
            
            \addplot[thick, mark=+, mark size=1pt, color=blue] table [x index=1, y index=3,
            header=true, col sep = comma]
            {./figuredata/SINGLE_MACHINE/ENRON/ESVI.1P16.K128.txt};
            \addlegendentry{ESVI-TOPK C=128}
        \end{axis}
      \end{tikzpicture} 
       \begin{tikzpicture}[scale=0.6]
        \begin{axis}[xmode=log, minor tick num=1,
          legend style={at={(0.7,0.18)}, anchor= south west},
          title={Ny Times dataset: P=1$\times$16 K=64 C=\{1, 4, 16, 32, 64\}},
          xlabel={time}, ylabel={ELBO}]
            \addplot[thick, solid, mark=*, mark size=1pt, color=red] table [x index=1, y index=3,
            header=true, col sep = comma]
            {./figuredata/SINGLE_MACHINE/NYTIMES/TOPK.1P16.K64.C1.txt};
            \addlegendentry{ESVI-TOPK C=1}
            
            \addplot[thick, solid, mark=triangle*, mark size=1pt, color=green] table [x index=1, y index=3,
            header=true, col sep = comma]
            {./figuredata/SINGLE_MACHINE/NYTIMES/TOPK.1P16.K64.C4.txt};
            \addlegendentry{ESVI-TOPK C=4}
            
            \addplot[thick, solid, mark=square*, mark size=1pt, color=brown] table [x index=1, y index=3,
            header=true, col sep = comma]
            {./figuredata/SINGLE_MACHINE/NYTIMES/TOPK.1P16.K64.C16.txt};
            \addlegendentry{ESVI-TOPK C=16}
            
            \addplot[thick, solid, mark=diamond*, mark size=1pt, color=magenta] table [x index=1, y index=3,
            header=true, col sep = comma]
            {./figuredata/SINGLE_MACHINE/NYTIMES/TOPK.1P16.K64.C32.txt};
            \addlegendentry{ESVI-TOPK C=32}
            
            \addplot[thick, solid, mark=+, mark size=1pt, color=blue] table [x index=1, y index=3,
            header=true, col sep = comma]
	    {./figuredata/SINGLE_MACHINE/NYTIMES/ESVI.1P16.K64.txt};
            \addlegendentry{ESVI-TOPK C=64}
        \end{axis}
      \end{tikzpicture}
\caption{Effect of varying $C$ (cutoff for k) in ESVI-LDA-topk}
  \label{fig:exp_topk_scaling}
\end{figure}
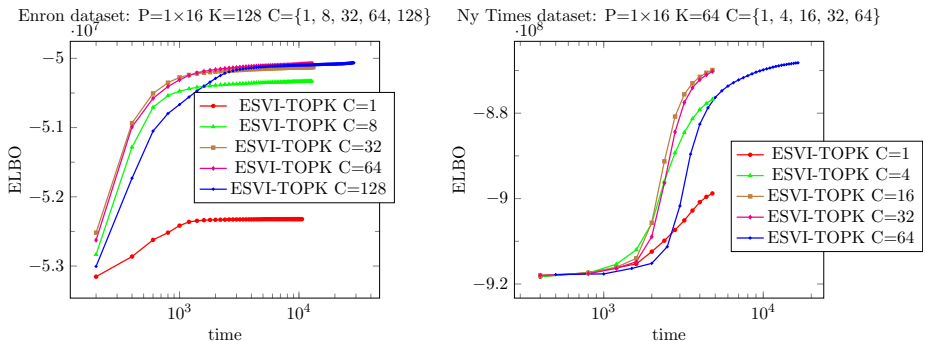

In this sub-section, we investigate the effect of varying the value of cutoff $C$ in the ESVI-LDA-TOPK method. 
While the approximation of evidence lower bound (ELBO) must get more accurate as $C \to K$, there 
might exist a choice of $C << K$, which still gives a reasonably good enough approximation. This will give us a significant 
boost in speed. Figure \ref{fig:exp_topk_scaling} shows the result of our experiment. On Enron dataset,
we varied $C$ as $1, 8, 32, 64, 128$ with the true $K = 128$ as our baseline. On NY Times dataset, we varied $C$ as $1, 4, 16, 32, 64$
with the true $K = 64$ as baseline. As we expected, setting the cutoff to a value too low leads to very slow convergence. However, it
is interesting to note that at a cut off value of roughly $\frac{K}{4}$ (32 on Enron and 16 on NY Times), we get a good result on par with 
using the baseline. On the larger datasets, PubMed and UMBC-3B also, we observed a similar behavior where setting $C=16$
was enough to achieve a similar ELBO as the baseline.
Figure \ref{fig:exp_topk_scaling} shows the results of this experiment.

\subsubsection{Fixing $C$ (cutoff for k) and scaling to large $K$ in ESVI-LDA-TOPK}
\label{subsec:fixc_varyk}
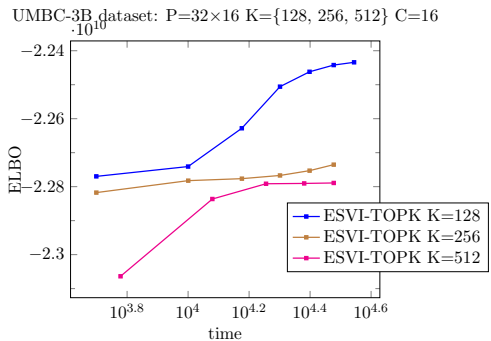
\begin{figure}[ht]
 \centering
      \begin{tikzpicture}[scale=0.6]
        \begin{axis}[xmode=log, minor tick num=1,
          legend style={at={(0.7,0.1)}, anchor= south west},
          title={UMBC-3B dataset: P=32$\times$16 K=\{128, 256, 512\} C=16},
          xlabel={time}, ylabel={ELBO}]
            \addplot[thick, solid, mark=square*, mark size=1pt, color=blue] table [x index=1, y index=3,
            header=true, col sep = comma]
            {./figuredata/MULTI_MACHINE/UMBC/TOPK.32P16.K128.C16.hard.txt};
            \addlegendentry{ESVI-TOPK K=128}

            \addplot[thick, solid, mark=square*, mark size=1pt, color=brown] table [x index=1, y index=3,
            header=true, col sep = comma]
            {./figuredata/MULTI_MACHINE/UMBC/TOPK.32P16.K256.C16.hard.txt};
            \addlegendentry{ESVI-TOPK K=256}
            
            \addplot[thick, solid, mark=square*, mark size=1pt, color=magenta] table [x index=1, y index=3,
            header=true, col sep = comma]
            {./figuredata/MULTI_MACHINE/UMBC/TOPK.32P16.K512.C16.hard.txt};
            \addlegendentry{ESVI-TOPK K=512}            

        \end{axis}
      \end{tikzpicture}
\caption{Effect of varying $K$ by fixing $C$ (cutoff for k) in ESVI-LDA-TOPK}
  \label{fig:exp_fixc_varyk}
\end{figure}

In this sub-section, we present results showing how, once we have picked a suitable cutoff $C$
for the our ESVI-LDA-TOPK method, we can scale our algorithm to very large number of topics such as
$K=256$ and $K=512$ on the largest dataset: UMBC-3B. Figure \ref{fig:exp_fixc_varyk} below demonstrates this.

\end{document}